\documentclass[sigconf]{acmart}
\AtBeginDocument{%
  }

\setcopyright{acmlicensed}
\copyrightyear{2026}
\acmYear{2026}
\setcopyright{cc}
\setcctype{by-nc-nd}

\acmConference[KDD 2026] {Proceedings of the 32nd ACM SIGKDD Conference on Knowledge Discovery and Data Mining V.2}{August 9--13, 2026}{Jeju Island, Republic of Korea.}

\acmBooktitle{Proceedings of the 32nd ACM SIGKDD Conference on Knowledge Discovery and Data Mining V.2 (KDD 2026), August 9--13, 2026, Jeju Island, Republic of Korea}

\acmISBN{979-8-4007-2259-2/2026/08}
\acmDOI{10.1145/3770855.3817949}

\usepackage{url}
\usepackage{enumitem}
\usepackage{multirow}
\usepackage[table]{xcolor}
\usepackage{subcaption}
\usepackage{algorithm}
\usepackage{algpseudocode}
\usepackage{dsfont}

\newcommand{{\method}}{EGSteal}

\begin{document}

\title{Do Explanations Increase the Risk of Decision Logic Leakage? Explanation-Guided Stealing of Graph Models}

\author{Bin Ma}
\email{bma662@connect.hkust-gz.edu.cn}
\affiliation{
  \institution{The Hong Kong University of Science and Technology (Guangzhou)}
  \city{Guangzhou}
  \country{China}
}

\author{Yuyuan Feng}
\email{30920231154331@stu.xmu.edu.cn}
\affiliation{
  \institution{Xiamen University}
  \city{Xiamen}
  \country{China}
}

\author{Minhua Lin}
\email{mfl5681@psu.edu}
\affiliation{
  \institution{The Pennsylvania State University}
  \city{State College}
  \country{USA}
}

\author{Enyan Dai}
\email{enyandai@hkust-gz.edu.cn}
\affiliation{
  \institution{The Hong Kong University of Science and Technology (Guangzhou)}
  \city{Guangzhou}
  \country{China}
}

\begin{abstract}
Graph Neural Networks (GNNs) have become essential tools for analyzing graph-structured data in domains such as drug discovery and financial analysis, leading to a growing demand for model transparency. Recent advances in explainable GNNs have addressed this need by revealing important subgraphs that influence predictions, but these explanation mechanisms may inadvertently expose these models to security risks. This paper investigates how such explanations potentially leak critical decision logic that can be exploited for model stealing. We propose {\method}, a novel stealing framework that integrates explanation alignment for capturing decision logic with guided data augmentation for efficient training under limited queries, enabling effective replication of both the predictive behavior and underlying reasoning patterns of target models. Experiments on molecular graph datasets demonstrate that our approach shows advantages over conventional methods in model stealing. This work highlights important security considerations for the deployment of explainable GNNs in sensitive domains and suggests the need for protective measures against explanation-based attacks. Our code is available at \url{https://github.com/beanmah/EGSteal}.
\end{abstract}



\maketitle

\section{Introduction}

\begin{figure}[t]
  \centering
  \includegraphics[width=0.95\linewidth]{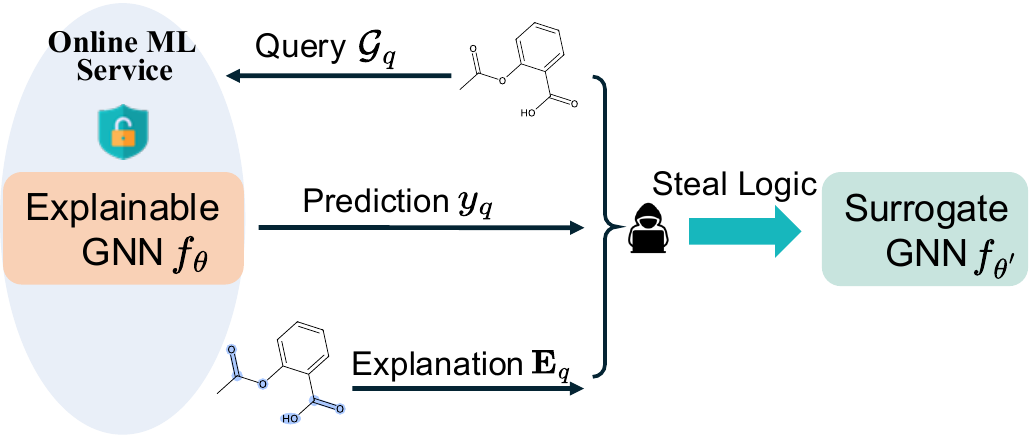}
  \caption{Explainable GNNs and the potential leakage of decision logic from explanations.}
  \vspace{-1em}
  \label{fig:intro}
\end{figure}

Graph Neural Networks (GNNs) have emerged as powerful tools for analyzing graph-structured data, demonstrating remarkable success across various critical scenarios including drug discovery~\cite{sun2020graph}, financial analysis~\cite{lv2019auto}, and social network analysis~\cite{hamilton2017inductive}.  The need for trust in these high-stakes scenarios has accelerated the development of explainable GNN methods~\cite{ying2019gnnexplainer,pope2019explainability,yuan2021explainability}. As illustrated in Fig.~\ref{fig:intro}, explainable GNNs generally identify important subgraphs as the explanation to reflect the decision logic on the input samples. 

Although explainable GNN mechanisms enhance the model transparency and interpretability, the provided explanations can inadvertently leak internal decision logic, thereby increasing the risk of model stealing. In practice, to protect intellectual property, GNN models in critical applications such as drug screening are typically deployed exclusively via online services. As shown in Fig.~\ref{fig:intro}, malicious attackers could query these deployed explainable GNNs to obtain explanations that potentially expose internal decision logic.  These explanations may enable attackers to construct a high-fidelity surrogate model that precisely replicates the decision logic of private models. Hence, this raises a concern about the vulnerability of explainable GNNs to model stealing attacks.

Previous studies have investigated model stealing attacks against vanilla GNNs, which do not provide explanations for their predictions. Wu \textit{et al.}~\cite{wu2022model} conduct the first investigation into GNN model stealing attacks. Specifically, they train the surrogate model to replicate the predictions of the target model on queried samples. More recent studies further explore strategies such as embedding alignment~\cite{shen2022model} and graph augmentation~\cite{podhajski2024efficient, guan2024realistic} to improve attack efficacy. However, these methods generally treat the target model as a black box, ignoring the additional information leaked through explanation mechanisms. Regarding the investigation of model leakage through explanations, early research in computer vision~\cite{milli2019model, yan2023explanation} demonstrated that explanations can reveal information about model parameters. However, the unique graph structures and message-passing mechanisms in GNNs significantly differentiate their decision-making and explanation processes. Thus, previous methods for images are not directly applicable to explainable GNNs. Consequently, it is crucial to specifically investigate the potential threat of model stealing attacks against explainable GNNs.

Two major questions remain to answer for the investigation of model stealing attacks against explainable GNNs. \textit{First}, can explanations provided by explainable GNNs facilitate the extraction of the target model's prediction logic?  While explanations could provide insights into the GNN's reasoning process, converting this information into effective training signals for the surrogate GNN is non-trivial. \textit{Second}, can attackers efficiently perform model stealing against explainable GNNs under limited query budgets? Real-world scenarios often impose strict constraints on the number of allowed queries, which requires methods to maximize the utility of each interaction with the target explainable GNN. Therefore, how to conduct effective model stealing attack against explainable GNNs within limited queries becomes another critical challenge.

To address these challenges comprehensively, we first propose a causal view of model stealing attacks with explanations. This causal analysis highlights the necessity of explanation alignment for effective logic stealing. In addition, the style invariance property identified in Sec.~\ref{sec:causal} indicates that predictions under various style interventions can be directly inferred without additional queries, thus enabling efficient augmentation of the surrogate model's training set. Motivated by this causal perspective, we propose an \underline{E}xplanation-guided \underline{G}NN \underline{steal}ing framework ({\method}). Specifically, {\method} employs a novel rank-based explanation alignment loss to ensure that the surrogate model not only replicates the predictive behavior of the target model but also captures its underlying decision logic. Furthermore, inspired by our causal analysis, {\method} incorporates explanation-guided data augmentation, which can facilitate effective model stealing under limited query budgets. Our {\method} verifies the vulnerability of explainable GNNs under model stealing attacks, which could inspire effective defense methods to enhance model security.
In summary, our main contributions are: 

\begin{itemize}[leftmargin=*]
    \item We focus on a novel problem of model stealing attacks on explainable GNNs. In addition, the model stealing with explanations is analyzed from a causal view, which motivates effective solutions.  
    \item We propose a new framework integrating explanation alignment and explanation-guided augmentation to effectively replicate the logic of explainable GNNs under limited query budgets.
    \item Extensive experiments on various GNN architectures and explanation methods demonstrate that our approach can effectively performs model stealing that captures both prediction behaviors and underlying decision logic.
\end{itemize}

\section{Related Works}

\noindent \textbf{Explainable Graph Neural Networks}.
GNNs have demonstrated significant success in many critical domains. To enhance the trustworthiness of GNN predictions, researchers have proposed various explanation methods in the form of important subgraphs. Early methods widely use gradient-based techniques~\cite{baldassarre2019explainability} to identify important features, while CAM~\cite{zhou2016learning, pope2019explainability} generates explanations by examining the final layer's feature maps. Grad-CAM~\cite{selvaraju2017grad, pope2019explainability} further incorporates gradient and CAM to better localize important regions. One notable advancement, GNNExplainer~\cite{ying2019gnnexplainer} formulates explanations as a compact subgraph maximizing the mutual information with the model's prediction. PGExplainer~\cite{luo2020parameterized} further accelerates this process by training a parameterized explanation generator. Other approaches, such as GraphLIME~\cite{huang2022graphlime} adapt local interpretable model-agnostic explanations, and SubgraphX~\cite{yuan2021explainability} employ Monte Carlo tree search. For more related works on explainable GNNs, please refer to Appendix~\ref{app:related_works1}.

\noindent \textbf{Model Stealing Attacks against GNNs}.
Model stealing or model extraction attacks aim to steal the target model information by training a similar surrogate model that matches its prediction behavior~\cite{dai2022comprehensive, tramer2016stealing, wu2022model}. Wu et al.~\cite{wu2022model} conducted the first investigation on  GNN model stealing attacks by training a surrogate model through queried APIs. Shen et al.~\cite{shen2022model} further extend the framework to inductive GNNs, which better reflect real-world scenarios. Building upon this direction, EfficientGNN~\cite{podhajski2024efficient} augments the shadow dataset with contrastive learning and spectral graph augmentations, and Zhuang et al.~\cite{zhuang2024unveiling} recently propose STEALGNN, introducing a more challenging scenario where attackers have no access to any real graph data. With the wide deployment of explainable models, some recent research in computer vision has investigated the risk of explainable models to model logic extraction. For instance, Milli et al.~\cite{milli2019model} demonstrate that gradient-based explanations of a model can reveal the model itself. 
Yan et al.~\cite{yan2023explanation} further introduce an extra CNN autoencoder to utilize the representations learned by reconstructing the explanations as data augmentations. In this study, we first address the vulnerability of explainable GNNs to model stealing attacks. Please refer to Appendix~\ref{app:related_works2} for more related works on GNN stealing attacks.

\section{Preliminaries}

Let $\mathcal{G} = (\mathcal{V}, \mathcal{E}, \mathbf{X})$ denote a graph, where $\mathcal{V}$ is the set of nodes, $\mathcal{E} \subseteq \mathcal{V} \times \mathcal{V}$ represents the edge set, and $\mathbf{X} \in \mathbb{R}^{|\mathcal{V}| \times d}$ is the node feature matrix with dimension $d$. Let $f_{\theta}$ denote the target model with parameters $\theta$ trained on $\mathcal{D}_T$. For a explainable target model, the explanation mechanism is represented by $\phi$. 
In a model stealing attack, the attacker has a shadow dataset containing a handful of query graphs $\mathcal{D}_Q$. Given the above notations, we introduce the preliminaries in the following.

\subsection{Preliminaries of GNN Explanations}
To integrate both node features and graph topology for representation learning, Graph Neural Networks (GNNs) generally adopt the message-passing mechanism to update node representations by aggregating the information from their neighborhood nodes. Specifically, the embedding of node $v$ at layer $k$ is updated as:
\begin{equation}
    \mathbf{h}_v^{(k)} = \text{AGGREGATE}(\mathbf{h}_v^{(k-1)}, \{\text{MSG}(\mathbf{h}_v^{(k-1)}, \mathbf{h}_u^{(k-1)})\}),
\end{equation}
where $u \in \mathcal{N}(v)$ denoting its neighborhoods. Various explanation methods have been proposed to enhance the explainability of GNNs. The majority of these methods providing explanations in the form of important subgraphs such as Graph-CAM~\cite{pope2019explainability} and GNNExplainer~\cite{ying2019gnnexplainer}. 
Therefore, we focus on subgraph explanations. Formally, given an input graph $\mathcal{G}=(\mathcal{V}, \mathcal{E})$ and a GNN  $f_{\theta}$, an explainer $\phi$ identifies a subgraph $\mathcal{G}_E$ as the explanation. In practice, explanations often take a soft form and can be uniformly represented as a node importance vector:
\begin{equation} 
    \mathbf{E} = [E_1, \dots, E_{|\mathcal{V}|}]^\top = \phi(f_{\theta}, \mathcal{G}) \in \mathbb{R}^{|\mathcal{V}|},
\end{equation}
where a higher value of $E_i$ indicates a greater contribution of node $v_i$ to the model's graph classification decision. For more details on GNN explanation methods, please refer to Appendix~\ref{appendix:explanation_methods}.

\subsection{Threat Model}
The goal of the adversary is to train a surrogate model that matches both the accuracy and decision boundary of the target model under a limited query budget. For the attacker's knowledge, our attack operates under the black-box setting where the target model's architecture, parameters, and internal explanation mechanism are unknown. The attacker possesses a shadow dataset $\mathcal{D}_S$ and a query budget $Q$ for interactions with the target model. For each query graph $\mathcal{G}_q \in \mathcal{D}_Q$, the attacker receives both the model prediction $\hat{y}_q $ and the explanation output $\mathbf{E}_q$, which is a node importance score vector.
For the shadow data, we consider two settings (i) In-distribution setting: where the shadow data $\mathcal{D}_S$ follows the same distribution as the target model's training data $\mathcal{D}_T$; (ii) Cross-distribution setting:  $\mathcal{D}_S$ and  $\mathcal{D}_T$ are not from the sample distribution but are in the same domain.

\section{Methodology}
\label{sec:methodology}

In this section, we first analyze the model stealing with explanations from a causal perspective. 
The key insight is that explanations partition an input graph into two distinct parts: a causal explanation subgraph $\mathcal{G}_E$ containing essential structures determining model predictions, and a non-causal style subgraph $\mathcal{G}_S$ whose modifications do not affect predictions. The identified causal mechanisms could guide surrogate model training for effective logic extraction. In addition, the causal analysis enables the generation of diverse training samples through interventions on non-causal components.

\subsection{Causal Analysis of Graph Model Stealing with Explanations}
\label{sec:causal}

\begin{figure}[t]
  \centering
  \includegraphics[width=0.35\linewidth]{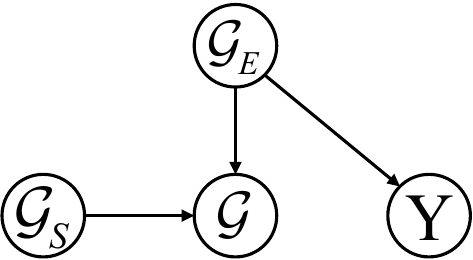}
  \caption{SCM of explainable graph prediction.}
  \label{fig:scm}
\end{figure}

We adopt a causal perspective to analyze the model stealing with explanations \cite{chen2022learning, wu2022discovering, sui2022causal}. Specifically, we propose a Structural Causal Model (SCM)~\cite{pearl2009causality} as illustrated in Figure~\ref{fig:scm}. The SCM includes four key variables: the explanation subgraph $\mathcal{G}_E$, the style subgraph $\mathcal{G}_S$, the input graph $\mathcal{G}$, and the prediction outcome $Y$. Each directed link represents a causal relationship between the variables. Specifically, we detail the causal relationships as follows: 
\begin{enumerate} [leftmargin=*]
    \item $\mathcal{G}_E \rightarrow \mathcal{G} \leftarrow \mathcal{G}_S$: The graph $\mathcal{G}$ is formed of the explanation subgraph $\mathcal{G}_E$ and the style subgraph $\mathcal{G}_S$.
    \item $\mathcal{G}_E \rightarrow Y$: The explanation subgraph $\mathcal{G}_E$ consists of critical decision-related structures, which determine the prediction $Y$, while the style subgraph $\mathcal{G}_S$ has no causal effect on $Y$.
\end{enumerate}

Following the principle of Independence of Mechanisms, interventions on the style variable $\mathcal{G}_S$ should not change the conditional distribution $p(Y|\mathcal{G}_E)$. Formally, for the target model, the following invariance property holds:
\begin{equation}
    p^{do(\mathcal{G}_S = \mathcal{G}_S^i)}(Y|\mathcal{G}_E;\theta)=p(Y|\mathcal{G}_E;\theta), \quad \forall \mathcal{G}_S^i \in \mathbb{G}_S,
    \label{eq:causal_surrogate}
\end{equation}
where $p^{do(\mathcal{G}_S = \mathcal{G}_S^i)}$ denotes the distribution obtained by intervening to set the style subgraph variable $\mathcal{G}_S$ to a particular value $\mathcal{G}_S^i$, and $\mathbb{G}_S$ represents the set of all possible style subgraph variations.
Recall that the objective of model stealing is to learn a surrogate model $f_{\theta'}$ that simulate the behaviors of the target model. Given Eq.(\ref{eq:causal_surrogate}), the model stealing with explanations should further meet:
\begin{equation}
\begin{split}
    p^{do(\mathcal{G}_S = \mathcal{G}_S^i)}(Y|\mathcal{G}_E;\theta') &= p^{do(\mathcal{G}_S = \mathcal{G}_S^i)}(Y|\mathcal{G}_E;\theta) \\
    &= p(Y|\mathcal{G}_E;\theta), \quad \forall \mathcal{G}_S^i \in \mathbb{G}_S.
\end{split}
\label{eq:causal_obj}
\end{equation}
To satisfy the principle in Eq.(\ref{eq:causal_obj}), the surrogate model $f_{\theta'}$ should fulfill the following two properties:
\begin{itemize}[leftmargin=*]
    \item \textbf{Explanation Alignment}: The explanation graphs generated by the surrogate model $f_{\theta'}$ should align with those of the target model $f_{\theta}$. This requires explicit alignment of two models.
    \item \textbf{Style Invariance}:  
    Predictions of the surrogate model should remain invariant under style interventions. Since $ p(Y|\mathcal{G}_E;\theta)$ can be obtained through a single query to the target model, predictions under various style interventions can be determined without additional queries. This enables augmentation of the surrogate model's training set without extra query costs.
\end{itemize}

\subsection{Framework of {\method}}

\begin{figure}[t]
  \centering
  \includegraphics[width=0.99\linewidth]{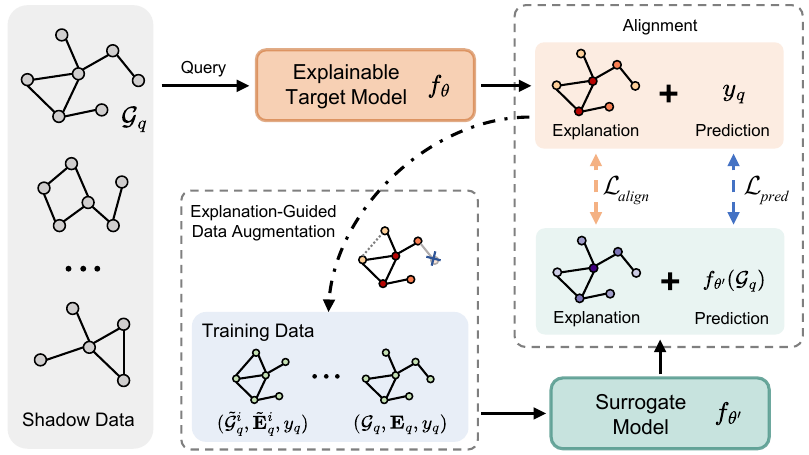}
  \caption{Overview framework of our \method.}
  \label{fig:pipeline}
\end{figure}

In this subsection, we elaborate on {\method}, which is designed to satisfy the properties derived from our causal analysis of model stealing with explanations. As illustrated in Fig.~\ref{fig:pipeline}, our proposed framework comprises two major components: the {logic stealing via explanation alignment} and the {explanation-guided data augmentation}. In particular, motivated by the principle of explanation alignment, we introduce a novel rank-based explanation alignment loss, explicitly ensuring that the surrogate model not only replicates the predictive behavior of the target model but also faithfully captures its underlying reasoning patterns. Moreover, guided by the principle of style invariance, {\method} deploys an explanation-guided data augmentation to perform style interventions. This allows diverse training data generation without additional queries, which enables effective model stealing with limited query budgets.

\subsubsection{Logic Stealing via Explanation Alignment}
Model stealing requires to capture the reasoning process of target graph neural network. The necessity of aligning explanations is further justified by the causal analysis presented in Sec.~\ref{sec:causal}. Therefore, we propose an explanation alignment module detailed below.

\noindent \textbf{Explanation Mechanism for Surrogate Model}.
To enable explanation alignment, both target and surrogate models must have interpretable explanation mechanisms. While various explanation methods, such as GNNExplainer~\cite{ying2019gnnexplainer}, could  serve this purpose, we specifically adopt CAM~\cite{baldassarre2019explainability, pope2019explainability} for two key reasons: 
\begin{itemize} [leftmargin=*]
    \item CAM obtains node importance scores without requiring additional training of explainers, thereby efficiently and faithfully reflecting the surrogate model’s decision logic; 
    \item Explanations generated by CAM naturally support gradient-based optimization, allowing direct backpropagation of explanation alignment losses into the surrogate model's parameters. 
\end{itemize}
CAM aims to interpret GNN predictions by identifying nodes that contribute significantly to classifications. It leverages node-level embeddings produced by the GNN encoder and traces the prediction process back to these embeddings using classifier weights. Formally, given an input graph with node features $\mathbf{X}$ and adjacency matrix $\mathbf{A}$, the GNN encoder computes node-level embeddings $F_{k,v}(\mathbf{X}, \mathbf{A})$, where $k$ indexes feature dimensions and $v$ indexes nodes. Then the node importance score of node $v$ in the surrogate model $f_{\theta'}$ are given by:
\begin{equation}
   E_{v}' = \sum_{k} w^{c}_{k} \cdot F_{k,v}(\mathbf{X},\mathbf{A}),
    \label{eq:cam3}
\end{equation}
where $w_k^c$ represents the classifier weight for feature dimension $k$ and the predicted class $c$.
This approach backpropagates class-specific weights to the pre-pooling node features, revealing each node's contribution to the final prediction. Nodes with higher importance scores thus have a stronger influence on the model’s decision for the corresponding class. 

\noindent \textbf{Rank-Based Explanation Alignment Loss}.
Given the explanations from both surrogate and target models, a natural approach to explanation alignment is to directly minimize the difference between their node importance scores, for instance by applying Mean Squared Error (MSE) loss. However, such a straightforward approach overlooks a key challenge: raw importance scores produced by different models or explanation methods often vary widely in scale and distribution, limiting the effectiveness of direct alignment.
To address this issue, we propose aligning explanations based on relative node rankings rather than absolute importance scores. The rationale is that, despite variations in magnitude, the relative ordering of node importance tends to remain stable and effectively reflects the underlying reasoning process. 
Following this intuition, we propose a rank-based explanation alignment loss inspired by RankNet~\cite{burges2005learning}. For any pair of nodes $(i,j)$ in a graph, the importance ranking given by the target model is denoted as $r_{ij}$, where $r_{ij} = 1$ if node $i$ is ranked higher (more important) than node $j$ according to the target model’s explanations, and $r_{ij} = 0$ otherwise. We compute the difference of their importance scores in the surrogate model as $\Delta_{ij} = E_i' - E_j'$. Formally, the rank-based explanation alignment loss can be computed by:
\begin{equation}
  \mathcal{R}(\Delta_{ij}, r_{ij}) = -r_{ij} \log \sigma(\Delta_{ij}) - (1 - r_{ij}) \log(1 - \sigma(\Delta_{ij})),
  \label{eq:rank2}
\end{equation}
where $\sigma$ is the sigmoid function. To account for all pairwise relationships in the graph, we define the final alignment loss as the average ranking loss over all node pairs:
\begin{equation}
  l_{\text{align}} = \frac{1}{|\mathcal{P}|} \sum_{(i,j) \in \mathcal{P}} \mathcal{R}(\Delta_{ij}, r_{ij}),
  \label{eq:align}
\end{equation}
where $\mathcal{P}$ denotes the set of all node pairs with $i < j$ to avoid duplicate comparisons. By explicitly enforcing node-ranking consistency, this alignment ensures that the surrogate model robustly captures the internal decision logic of the target model.

\subsubsection{Explanation-Guided Data Augmentation}

Recall from the causal analysis of model stealing with explanations (Eq.~\ref{eq:causal_obj}) that the surrogate model should satisfy the following style invariance property: $p^{do(\mathcal{G}_S = \mathcal{G}_S')}(Y|\mathcal{G}_E;\theta') = p(Y|\mathcal{G}_E;\theta)$, where $\mathcal{G}_S' \in \mathbb{G}_S$ represents any modified style subgraph. This property implies that predictions should remain invariant for samples generated through interventions on style subgraphs. We can directly infer labels of intervened samples without additional queries to the target model. Since we do not have access $\mathbb{G}_S$ , to simulate style variability, we conduct explanation-guided data augmentations as interventions on the style graph. Specifically, given a graph $\mathcal{G}$ and node importance scores $\mathbf{E}$ obtained from the target explainable GNN $f_{\theta}$, we first identify nodes with low importance as belonging to the non-causal style subgraph:
\begin{equation}
    \mathcal{V}_{S} = \{ v_i \in V \mid \text{rank}(\mathbf{E}_{i}) \leq \alpha \cdot |\mathcal{V}| \},
\end{equation}
where $\mathbf{E}_{i}$ denotes the importance score of node $i$, and $\alpha \in [0,1]$ is the proportion of important nodes.  We then perform the following explanation-guided data augmentation strategies:
\begin{itemize}[leftmargin=*]
    \item \textbf{Style Graph Node Dropping}: It randomly removes nodes in $\mathcal{V}_S$ to produce an augmented graph: $\tilde{\mathcal{G}} = \mathcal{G} \setminus \text{RandomSelect}(\mathcal{V}_{S}, \beta \cdot |\mathcal{V}_{S}|),$ where $\text{RandomSelect}(\mathcal{V}_{S}, \beta \cdot |\mathcal{V}_{S}|)$ denotes randomly selecting a fraction $\beta \in [0,1]$ of nodes from the identified style node set $\mathcal{V}_{S}$ during the augmentation process.
    \item \textbf{Style Graph Edge Perturbation}: This strategy randomly modifies edges connecting the nodes in $\mathcal{V}_{S}$. It will create intervened graph $\mathcal{\tilde{G}}$ with diverse style variations while preserving the causal structure represented by the high-importance nodes.
\end{itemize} 

Using the above explanation-guided data augmentation strategies, each queried graph $\mathcal{G}_q \in \mathcal{D}_Q$ can be augmented into multiple intervened graphs. As for each queried tuple $(\mathcal{G}_q, \mathbf{E}_q, y_q)$ from the target model, it can be augmented into $\{(\tilde{\mathcal{G}}_q^i, \tilde{\mathbf{E}}_q^i, \tilde{y}_q)\}_{i=1}^{K}$. 
Specifically, the prediction and explanation of each augmented sample $(\tilde{\mathcal{G}}_q^i, \tilde{\mathbf{E}}_q^i, y_q)$ are derived as follows:
\begin{equation}
    \tilde{y}_{q}^i = y_q,\quad
\tilde{\mathbf{E}}_q^i = \mathbf{E}_q[\tilde{\mathcal{G}}_q^i],
\label{eq:augment_tuple}
\end{equation}
where $\mathbf{E}_q[\tilde{\mathcal{G}}_q^i]$ represents the subset of the original explanation vector obtained by directly omitting nodes removed from the original graph $\mathcal{G}_q$ to construct $\mathcal{\tilde{G}}_q^i$. As suggested in Eq.(\ref{eq:augment_tuple}), the prediction $\tilde{y}_q$ on $\tilde{\mathcal{G}}_q^i$ remains $y_q$ for each augmented graph, which in line with the style invariance property. For the corresponding explanation $\tilde{\mathbf{E}}_q^i$, we directly omit nodes removed in $\tilde{\mathcal{G}}_q^i$. This is justified by the assumption that modifications of non-causal style graphs do not alter the relative importance rankings of the remaining causal nodes.

\subsubsection{Final Objective Function}

Our final objective function combines explanation alignment and explanation-guided data augmentation. Let $\mathcal{D}_Q$ and $\mathcal{D}_A$ denote the queried dataset and augmented dataset, respectively, the final objective function of {\method} of optimizing the surrogate model's parameters $\theta'$ can be written as:
\begin{equation}
    \min_{\theta'} \mathcal{L}_{\text{total}} = \sum_{\mathcal{G}_i \in \mathcal{D}_Q\cup \mathcal{D}_A} l_{\text{pred}}(y_i,f_{\theta'}(\mathcal{G}_i)) + \lambda l_{\text{align}}(\mathbf{E}_i, \mathbf{E}_i'),
    \label{eq:total}
\end{equation}
where $l_{\text{pred}}$ is the cross-entropy loss, $l_{\text{align}}$ is the rank-based explanation alignment loss in Eq.~\ref{eq:align} that aligns the node importance vectors $\mathbf{E}_i$ from the target model and $\mathbf{E}_i'$ from the surrogate model, and $\lambda$ balances the contributions of two loss terms.

\begin{algorithm}[t]
\caption{Explanation-Guided GNN Stealing ({\method})}
\label{alg:training}
\begin{algorithmic}[1]
\Require Target model $f_{\theta}$, shadow dataset $\mathcal{D}_s$, query budget $Q$, augmentation factor $K \in \mathbb{Z}^{+}$, augmentation prob. $\rho \in [0,1]$.
\Ensure Trained surrogate model $f_{\theta'}$.

\State \textit{// Phase 1: Data Collection \& Augmentation}
\State Sample a query set $\mathcal{D}_Q = \{\mathcal{G}_q\}_{q=1}^Q$ from $\mathcal{D}_s$
\State $\mathcal{D}_{\text{train}} \leftarrow \emptyset$
\For{each $\mathcal{G}_q \in \mathcal{D}_Q$}
    \State $(y_q, \mathbf{E}_q) \leftarrow \text{Query } f_\theta(\mathcal{G}_q)$
    \State $\mathcal{D}_{\text{train}} \leftarrow \mathcal{D}_{\text{train}} \cup \{(\mathcal{G}_q, \mathbf{E}_q, y_q)\}$
    \If{$\text{rand}(0,1) < \rho$}
        \State Generate $K$ samples $\{(\tilde{\mathcal{G}}_q^i, \tilde{\mathbf{E}}_q^i, y_q)\}_{i=1}^K$ via Eq.~\eqref{eq:augment_tuple}
        \State $\mathcal{D}_{\text{train}} \leftarrow \mathcal{D}_{\text{train}} \cup \{(\tilde{\mathcal{G}}_q^i, \tilde{\mathbf{E}}_q^i, y_q)\}_{i=1}^K$
    \EndIf
\EndFor

\State \textit{// Phase 2: Surrogate Model Training}
\State Initialize surrogate model $f_{\theta'}$ with random parameters $\theta'$
\Repeat
    \State Sample a mini-batch $\mathcal{B} \subset \mathcal{D}_{\text{train}}$
    \State Compute total loss $\mathcal{L}_{\text{total}}$ on $\mathcal{B}$ according to Eq.~\eqref{eq:total}
    \State Update $\theta'$ to minimize $\mathcal{L}_{\text{total}}$ via a gradient-based optimizer
\Until{converged}
\State \Return $f_{\theta'}$
\end{algorithmic}
\end{algorithm}

\subsubsection{Overall Training Procedure}

We present the complete training procedure for our explanation-guided model stealing framework in Algorithm~\ref{alg:training}. The algorithm consists of two main phases: 
(i) \textit{Data Collection \& Augmentation}, where we query the target model $f_\theta$ to obtain predictions and explanations, followed by generating style-intervened variants with probability $\rho$ to enrich the training set; and 
(ii)  \textit{Surrogate Model Training}, where the surrogate model $f_{\theta'}$ is optimized to minimize the unified objective $\mathcal{L}_{\text{total}}$ as defined in Eq.~\eqref{eq:total}. The computational complexity analysis can be found in Appendix~\ref{app:complexity} and \ref{appendix:training_time}. In practice, we implement the optimization process in Algorithm~\ref{alg:training} using the Adam optimizer with a learning rate of $10^{-3}$. We set the default augmentation factor $K=1$, generating one augmented variant for each selected query.

\section{Experiments}
\label{sec:experiments}

In this section, we conduct extensive experiments to answer the following research questions:

\begin{itemize}[leftmargin=*]
    \item \textbf{RQ1} How effective is {\method} at extracting both prediction behavior and underlying decision logic from explainable GNNs compared to existing model stealing methods?
    \item \textbf{RQ2} How does {\method} perform under different practical scenarios including varying query budgets, distribution shifts between shadow and target datasets, and low explanation qualities?
    \item \textbf{RQ3} How well does {\method} generalize across different node classification tasks, model architectures, and GNN explainers?
\end{itemize}

\begin{table*}[t]
  \caption{Performance of stealing target models trained from scratch.}
  \vspace{-0.5em}
  \label{tab:main_results_without_pretraining}
  \centering
  \resizebox{\linewidth}{!}{
  \begin{tabular}{llccccccccc}
    \toprule
    Dataset & Metric (\%) & Target & TS & MEA-GNN & GNNStealing & EfficientGNN & MRME & DET & STEALGNN & Ours \\
    \midrule
    \multirow{3}{*}{NCI1}
    & AUC    & $81.70$ & $74.13 \pm 2.57$ & $72.29 \pm 1.00$ & \underline{$77.42 \pm 2.01$} & $75.93 \pm 1.32$ & $74.25 \pm 2.50$ & $71.49 \pm 1.12$ & $69.07 \pm 2.75$ & \textbf{80.74 $\pm$ 1.18} \\
    & Fidelity    & \multicolumn{1}{c}{--} & $76.23 \pm 1.86$ & $78.71 \pm 1.45$ & \underline{$80.41 \pm 1.16$} & $77.42 \pm 2.94$ & $76.55 \pm 1.06$ & $71.31 \pm 1.43$ & $66.27 \pm 2.33$ & \textbf{87.78 $\pm$ 0.54} \\
    & Rank Corr. & \multicolumn{1}{c}{--} & $15.22 \pm 1.09$ & $11.37  \pm 1.05$ & \underline{$18.62 \pm 0.53$} & $14.90 \pm 1.23$ & $15.35 \pm 0.81$ & $9.60  \pm 1.30$ & $8.82 \pm 1.68$ & \textbf{42.38 $\pm$ 1.01} \\
    \midrule
    \multirow{3}{*}{NCI109}
    & AUC         & $80.20$ & $71.13 \pm 3.54$ & $71.11 \pm 1.07$ & \underline{$75.22 \pm 0.83$} & $73.94 \pm 0.87$ & $72.16 \pm 2.05$ & $71.19 \pm 0.83$& $ 70.24 \pm 0.90$ & \textbf{78.08 $\pm$ 0.87} \\
    & Fidelity    & \multicolumn{1}{c}{--} & $74.40 \pm 5.00$ & $69.02 \pm 5.13$ & \underline{$76.70 \pm 3.90$} & $76.51 \pm 2.82$ & $75.73 \pm 3.58$ & $70.08 \pm 2.15$ & $69.26 \pm 2.23$ & \textbf{85.99 $\pm$ 1.20} \\
    & Rank Corr.  & \multicolumn{1}{c}{--} & $12.91 \pm 1.69$ & $9.56 \pm 0.85$  & \underline{$14.93 \pm 1.65$} & $13.91 \pm 1.30$ & $13.71 \pm 1.11$ & $8.29 \pm 1.26$ & $10.27 \pm 1.16$ & \textbf{35.58 $\pm$ 0.33} \\
    \midrule
    \multirow{3}{*}{AIDS}
    & AUC         & $94.19$ & $89.77 \pm 1.10$ & $88.15 \pm 1.29$ & \underline{$90.66 \pm 1.56$} & $89.34 \pm 1.85$ & $89.79 \pm 1.10$ & $85.49 \pm 2.34$ & $84.17 \pm 1.41$ & \textbf{93.38 $\pm$ 0.54} \\
    & Fidelity    & \multicolumn{1}{c}{--} & $87.65 \pm 1.69$ & \underline{$88.25 \pm 2.43$} & $86.70 \pm 1.89$ & $87.00 \pm 2.85$ & $87.60 \pm 1.70$ & $83.95 \pm 1.71$ & $79.70 \pm 1.98$ & \textbf{93.25 $\pm$ 0.52} \\
    & Rank Corr.  & \multicolumn{1}{c}{--} & $35.60 \pm 1.86$ & $23.97 \pm 2.37$ & $35.75 \pm 2.57$ & \underline{$36.60 \pm 2.07$} & $35.67 \pm 1.84$ & $13.31 \pm 6.06$ & $30.68 \pm 1.82$ & \textbf{62.91 $\pm$ 0.91} \\
    \midrule
    \multirow{3}{*}{Mutagenicity}
    & AUC         & $82.57$ & $79.21 \pm 1.95$ & $79.80 \pm 0.78$ & \underline{$81.72 \pm 0.41$} & $79.38 \pm 0.94$ & $79.32 \pm 2.03$ & $79.86 \pm 1.99$ & $81.17 \pm 1.14$ & \textbf{82.20 $\pm$ 1.24} \\
    & Fidelity    & \multicolumn{1}{c}{--} & $80.67 \pm 3.32$ & $80.25 \pm 1.15$ & \underline{$85.05 \pm 1.19$} & $81.43 \pm 1.30$ & $80.81 \pm 3.27$ & $80.14 \pm 1.69$ & $77.51 \pm 1.54$ & \textbf{88.81 $\pm$ 0.88} \\
    & Rank Corr.  & \multicolumn{1}{c}{--} & $22.47 \pm 3.75$ & $21.25 \pm 1.81$ & \underline{$29.33 \pm 1.75$} & $21.76 \pm 1.05$ & $22.62 \pm 3.52$ & $21.85 \pm 2.38$ & $22.08 \pm 1.48$ & \textbf{44.96 $\pm$ 0.41} \\
    \bottomrule
  \end{tabular}
  }
\end{table*}

\begin{table*}[t]
  \caption{Performance of stealing pre-trained target models.}
  \vspace{-0.5em}
  \label{tab:main_results_with_pre-training}
  \centering
  \resizebox{\linewidth}{!}{
  \begin{tabular}{llccccccccc}
    \toprule
    Dataset & Metric (\%) 
      & Target & TS & MEA-GNN & GNNStealing & EfficientGNN & MRME & DET & STEALGNN & Ours \\ 
    \midrule
    \multirow{3}{*}{HIV} 
    & AUC       
      & $78.94$ & $59.71\pm3.43$ & $61.66\pm2.80$ & $60.25\pm3.11$ & \underline{$62.69\pm1.56$} & $61.73\pm1.73$ & $56.30\pm2.47$ & $58.39\pm3.89$ & \textbf{66.00 $\pm$ 1.76} \\
    & Fidelity  
      & \multicolumn{1}{c}{--} & $95.04\pm2.35$ & $97.34\pm0.34$ & \textbf{97.56 $\pm$ 0.00} & $97.19\pm0.40$ & $96.02\pm1.44$ & \underline{$97.55\pm0.01$} & $97.54\pm0.02$ & $95.54\pm1.26$ \\
    & Rank Corr. 
      & \multicolumn{1}{c}{--}  & $2.54\pm7.54$ & $2.89\pm4.62$ & \underline{$15.93\pm4.39$} & $2.70\pm2.61$ & $8.14\pm3.12$ & $-9.26\pm8.69$ & $-8.32 \pm 24.92$ & \textbf{37.42 $\pm$ 1.21} \\
    \midrule
    \multirow{3}{*}{Tox21}
    & AUC       
      & $83.69$ & \underline{$75.57\pm1.27$} & $70.17\pm0.60$ & $75.06\pm1.29$ & $74.47\pm1.45$ & $75.16\pm3.85$ & $71.46\pm1.34$ & $71.44\pm4.12$ & \textbf{76.30 $\pm$ 1.88} \\
    & Fidelity  
      & \multicolumn{1}{c}{--} & $86.22\pm1.88$ & $84.95\pm5.96$ & \underline{89.28$\pm$0.03} & $81.80\pm8.34$ & $86.87\pm2.44$ & $89.26\pm0.24$ & \textbf{89.39 $\pm$ 0.02} & $88.53\pm1.02$ \\
    & Rank Corr. 
      & \multicolumn{1}{c}{--} & $4.15\pm5.74$ & $-4.04\pm5.11$ & $11.95\pm3.15$ & $2.75\pm3.15$ & $10.52\pm4.65$ & $-21.19\pm2.20$ & \underline{$19.53\pm5.31$} & \textbf{45.01 $\pm$ 3.35} \\
    \midrule
    \multirow{3}{*}{BACE}
    & AUC       
      & $87.08$ & $66.37\pm5.73$ & \underline{$71.68\pm4.09$} & $71.12\pm6.77$ & $69.94\pm3.90$ & $70.34\pm3.88$ & $63.70\pm4.09$ & $57.70\pm2.07$ & \textbf{74.15 $\pm$ 3.02} \\
    & Fidelity  
      & \multicolumn{1}{c}{--} & $58.67\pm5.07$ & $60.66\pm8.65$ & $60.66\pm6.13$ & $60.79\pm8.48$ & \underline{$62.38\pm3.09$} & $56.07\pm4.93$ & $49.57\pm2.10$ & \textbf{64.88 $\pm$ 8.61} \\
    & Rank Corr. 
      & \multicolumn{1}{c}{--} & $0.53\pm0.87$ & $1.73\pm1.12$ & $3.47\pm4.80$ & $-0.74\pm3.77$ & \underline{$4.09\pm4.00$} & $-1.77\pm1.95$ & $-4.00\pm2.27$ & \textbf{12.42 $\pm$ 2.81} \\
    \bottomrule
  \end{tabular}
  }
\end{table*}

\subsection{Experimental Setup}
\label{subsection:experimental_setups}

\subsubsection{Datasets, Metrics, and Implementation Details} We conduct experiments on seven molecular graph datasets~\cite{Morris+2020}~\cite{hu2019strategies}, including NCI1, NCI109, Mutagenicity, AIDS, ogbg-molhiv, Tox21, and BACE. Each dataset is split into three parts: 40\% for training the target model, 20\% as the test set, and 40\% as the shadow dataset, and the query data with different budgets is sampled from the shadow dataset. The target model uses CAM as the default explainer. We evaluate our framework using three metrics: (1) Area Under the ROC Curve (AUC) on the test set to assess predictive performance, (2) Prediction Fidelity, which measures the agreement between surrogate and target model predictions, and (3) Rank Correlation of explanations, which quantifies the alignment of decision logic through Kendall's tau coefficient between explanations from both models. Metric formulations and implementation details are provided in Appendix~\ref{appendix:metrics} and~\ref{appendix:hyperparameters_and_training_details}.

\subsubsection{Compared Methods} We compare our Explanation-Guided GNN Extraction Attack with three categories of methods: (i) the fundamental \textbf{teacher-student (TS)} model extraction baseline via prediction APIs \cite{tramer2016stealing}; (ii) recent works on model stealing attacks targeting GNNs, including \textbf{MEA-GNN} \cite{wu2022model}, \textbf{GNNStealing} \cite{shen2022model}, \textbf{EfficientGNN} \cite{podhajski2024efficient}, and \textbf{STEALGNN} \cite{zhuang2024unveiling}; (iii) methods from other domains that investigate explanation-guided model extraction attacks, specifically \textbf{MRME} \cite{milli2019model} and \textbf{DET} \cite{yan2023explanation}. More details on these methods can be found in Appendix~\ref{appendix:compared_methods}.

\subsection{Model Stealing Performance}
\label{subsection:model_extraction_performance}

To address \textbf{RQ1}, we evaluate our {\method} against baseline methods across two experimental settings, as shown in Tables~\ref{tab:main_results_without_pretraining} and~\ref{tab:main_results_with_pre-training}.

\subsubsection{Stealing GNN Models Trained from Scratch}
Table~\ref{tab:main_results_without_pretraining} presents results on molecular datasets where GIN target models are trained from scratch. For all experiments in this setting, the budget for querying shadow data points is set to 30\% of the target model's training data size. Our method consistently outperforms all baselines across datasets, approaching the target model's performance with smaller gaps. A phenomenon emerges from the metrics: conventional approaches achieve reasonable prediction performance but struggle with explanation alignment. Our framework shows higher explanation correlation scores, with notable improvements over the strongest baselines. This suggests that explanation-guided learning helps the surrogate model better capture the target model's decision logic, rather than merely mimicking its output predictions. Additional results with GCN and GAT target models are presented in Appendix~\ref{app:additional_results_target_models}, showing consistent performance improvements on different target model architectures. To further validate the robustness of our approach, we analyze the performance under varying query budgets (10\% to 50\% of the target model's training data size, detailed in Appendix~\ref{app:query_budget_analysis}), which demonstrates that {\method} consistently outperforms baselines across different query ratios.

\subsubsection{Stealing Pre-trained GNN Models}
In Table~\ref{tab:main_results_with_pre-training}, we examine a more challenging scenario using more complex and powerful target models. These models are pre-trained following the strategy in \cite{hu2019strategies}, specifically combining graph-level multi-task supervised learning and node-level self-supervised learning (ContextPred), and then fine-tuned on downstream task datasets. For query data, the budget is set to 30\% of the target model's training data size for Tox21 and BACE, but only 10\% for HIV due to its larger scale. Tox21 is a multi-label classification dataset, from which we use only the "NR-AhR" label. Additionally, both HIV and Tox21 have class imbalance issues, creating additional extraction challenges. As shown in Table~\ref{tab:main_results_with_pre-training}, all methods achieve low overall performance compared to the target model. Nevertheless, our framework achieves significantly better performance, particularly in AUC and explanation correlation metrics, suggesting that incorporating the target model’s explanation information contributes to improved extraction.

\begin{figure*}[t]
    \centering
    \begin{subfigure}[b]{0.33\linewidth}
        \centering
        \includegraphics[width=\linewidth]{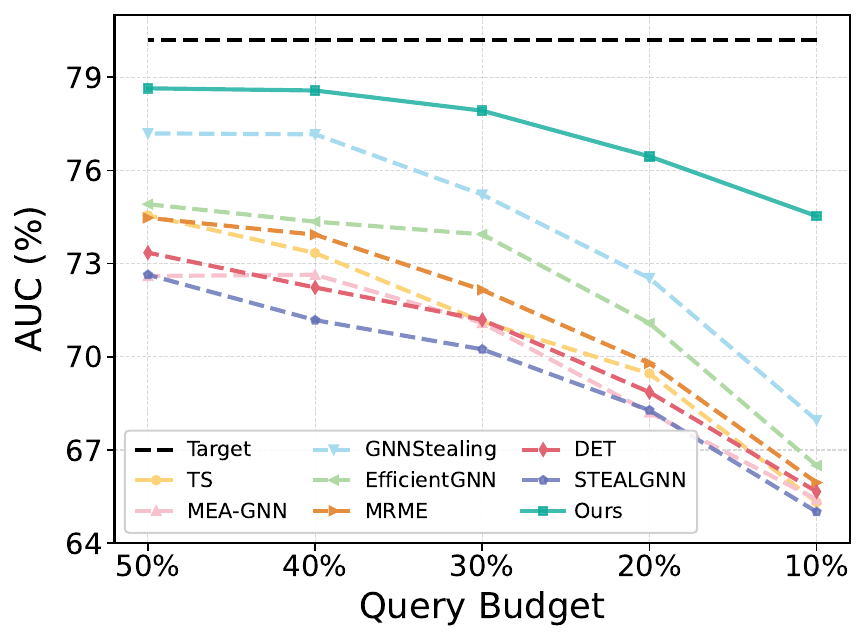}
        \caption{Resistance to Budget Decrease}
        \label{fig:query_budget_nci109}
    \end{subfigure}
    \hfill
    \begin{subfigure}[b]{0.36\linewidth}
        \centering
        \includegraphics[width=\linewidth]{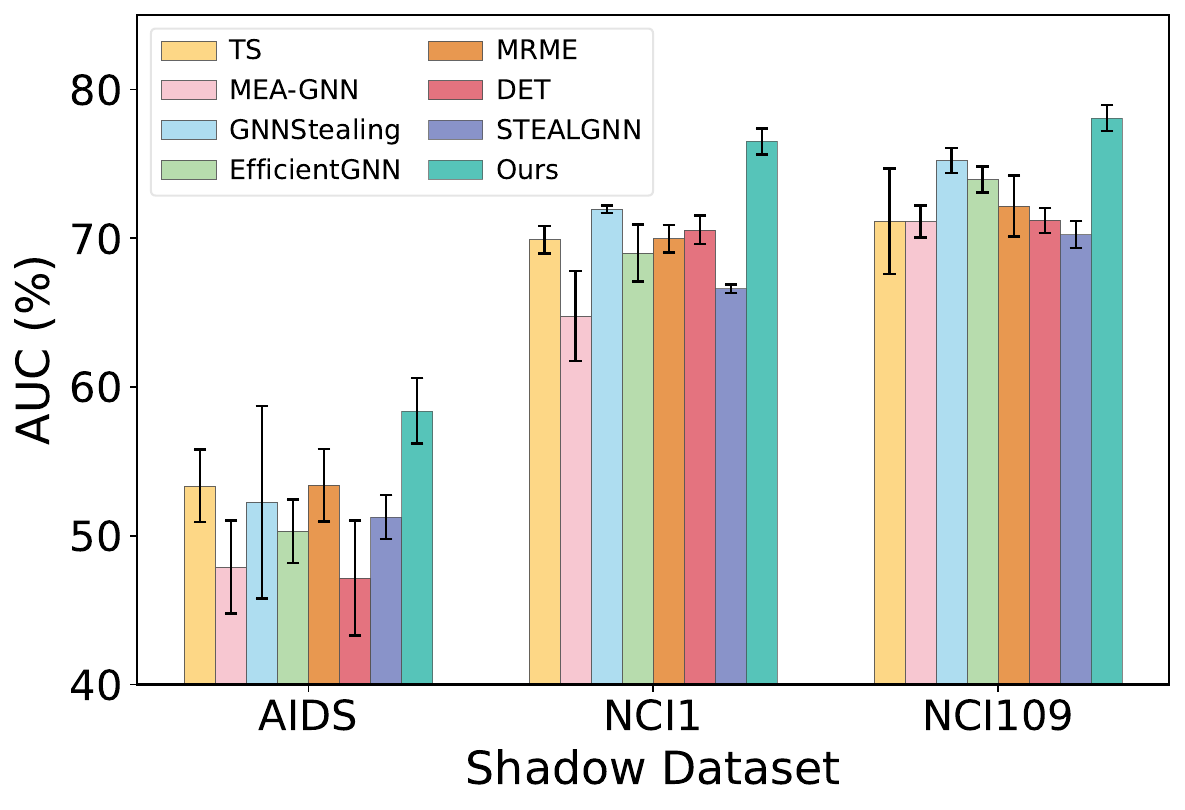}
        \caption{Cross-distribution robustness}
        \label{fig:cross_dist_sub}
    \end{subfigure}
    \hfill
    \begin{subfigure}[b]{0.28\linewidth}
        \centering
        \includegraphics[width=\linewidth]{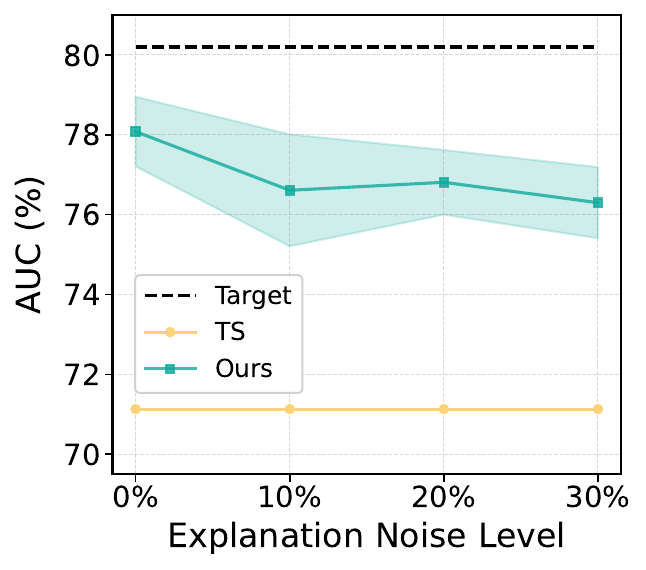}
        \caption{Noisy explanations}
        \label{fig:noise_impact_sub}
    \end{subfigure}
    \vspace{-0.5em}
    \caption{Robustness analysis under different practical scenarios. (a) Impact of query budget on extraction performance on NCI109 dataset. (b) Results when shadow dataset has a distribution shift from the target model's training set. (c) Performance under different levels of explanation noises on NCI109 dataset.}
    \label{fig:robustness_analysis}
\end{figure*}

\begin{table*}[t]
  \caption{Performance comparison on node classification tasks}
  \vspace{-0.5em}
  \label{tab:node_classification_task}
  \centering{
  \resizebox{\linewidth}{!}{
      \begin{tabular}{llccccccccc}
        \toprule
        Dataset & Metric (\%) & Target & TS & MEA-GNN & GNNStealing & EfficientGNN & MRME & DET & STEALGNN & Ours \\
        \midrule
        \multirow{3}{*}{PubMed}
        & AUC        & $90.42$ & $85.55 \pm 0.60$ & \underline{$86.10 \pm 0.63$} & $83.86 \pm 0.83$ & $85.93 \pm 0.78$ & $85.54 \pm 0.61$ & $75.38 \pm 3.65$ & $86.04 \pm 0.08$ &
        \textbf{87.51 $\pm$ 0.93} \\
        & Fidelity   & \multicolumn{1}{c}{--} & $82.54 \pm 3.02$ & $83.56 \pm 1.00$ & $75.54 \pm 0.64$ & $83.68 \pm 1.02$ & $83.06 \pm 2.36$ & $66.92 \pm 3.01$ & \underline{$86.16 \pm 0.73$} & \textbf{86.66 $\pm$ 0.51} \\
        & Rank Corr. & \multicolumn{1}{c}{--} & $33.29 \pm 1.40$ & $22.66 \pm 0.97$ & $12.82 \pm 2.66$ & $34.51 \pm 1.47$ & $33.30 \pm 1.46$ & $17.57 \pm 4.41$ & \underline{$36.00 \pm 0.59$} & \textbf{48.94 $\pm$ 0.87} \\
        \midrule
        \multirow{3}{*}{ogb-arxiv}
        & AUC        & $92.77$ & $87.04 \pm 0.93$ & \underline{88.83 $\pm$ 0.49} & $86.35 \pm 0.34$ & $86.95 \pm 0.55$ & $87.33 \pm 0.75$ & $82.58 \pm 1.40$ & $88.63 \pm 0.40$ & \textbf{89.76 $\pm$ 1.21} \\
        & Fidelity   & \multicolumn{1}{c}{--} & $61.35 \pm 2.45$ & \underline{$66.62 \pm 1.43$} & $62.92 \pm 1.88$ & $63.97 \pm 3.26$ & $63.01 \pm 1.66$ & $51.06 \pm 2.06$ &$66.06 \pm 0.85$ & \textbf{67.30 $\pm$ 2.10} \\
        & Rank Corr. & \multicolumn{1}{c}{--} & $26.21 \pm 3.54$ & $22.23 \pm 1.52$ & $2.99 \pm 2.81$ & $30.60 \pm 5.14$ & $26.44 \pm 3.54$ & \underline{$40.04 \pm 1.81$} & $36.50 \pm 3.25$ & \textbf{71.27 $\pm$ 0.70} \\
        \bottomrule
      \end{tabular}
      }
  }
\end{table*}

\subsection{Robustness Analysis}

To address \textbf{RQ2}, we evaluate our method under the practical scenarios of varying query budgets, distribution shift of shadow dataset, low quality explanations and explanations in binary mask. 

\subsubsection{Impact of Query Budget}
In real-world scenarios, query budgets are often limited due to cost, time, or detection avoidance considerations, making this a critical practical constraint. We investigate how reducing query budgets from 50\% to 10\% of the target model's training data size affects extraction performance. As shown in Figure~\ref{fig:query_budget_nci109}, our approach demonstrates more graceful degradation compared to baselines as the query budget decreases on NCI109. While all methods experience performance drops with fewer queries, our method maintains a more stable performance trajectory, with the performance gap over baselines widening at lower query ratios. This indicates that explanation information provides particularly valuable guidance when query access is limited. Results on additional datasets are provided in Appendix~\ref{app:query_budget_analysis}.

\subsubsection{Cross-Distribution Robustness}
We examine robustness to data distribution shifts by training the target model on NCI109, while using three different shadow datasets: NCI109 (same distribution), NCI1 (MMD=0.016), and AIDS (MMD=0.343). All models are tested on the NCI109 test set. As shown in Figure~\ref{fig:cross_dist_sub}, our approach maintains better performance than baselines as distribution divergence increases. While extraction performance decreases with increasing distribution shift for all methods, our explanation-guided approach demonstrates more consistent robustness. Notably, with the challenging AIDS dataset, our method achieves 6.68\% explanation correlation compared to TS's near-zero correlation (-0.77\%), indicating that explanations help capture transferable decision patterns beyond surface-level predictions. Implementation details and additional results and  are provided in Appendix~\ref{appendix:cross_distribution_implementation} and Appendix~\ref{appendix:cross_distribution_results}.

\begin{figure}[t]
  \centering
  \begin{subfigure}[b]{0.49\linewidth}
    \centering
    \includegraphics[width=\linewidth]{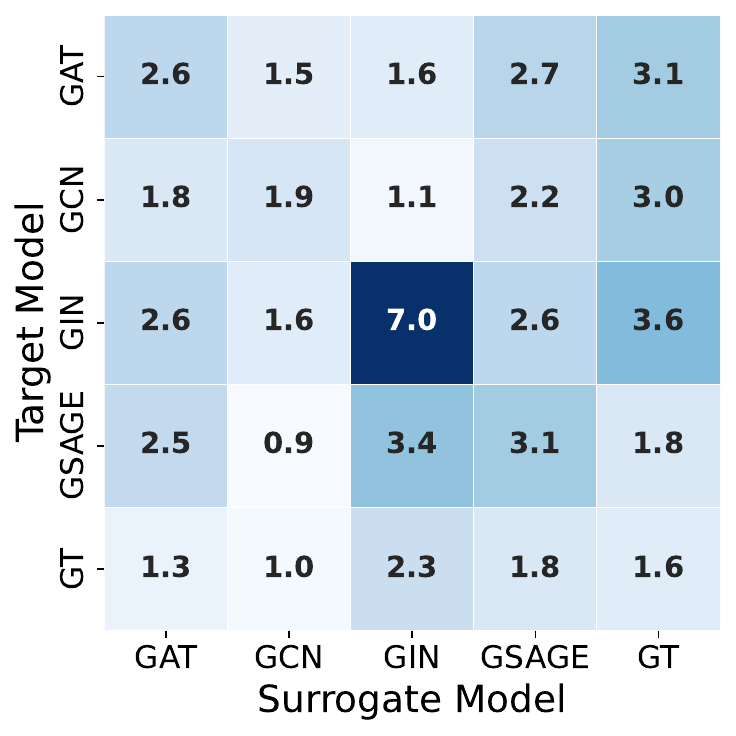}
    \caption{AUC gains (\%)}
  \end{subfigure}
  \hfill
  \begin{subfigure}[b]{0.49\linewidth}
    \centering
    \includegraphics[width=\linewidth]{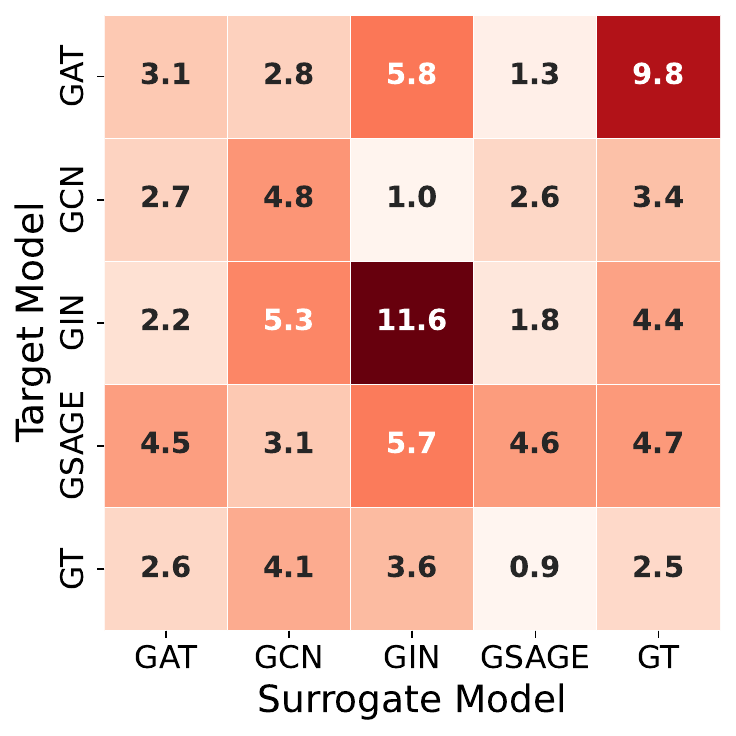}
    \caption{Fidelity gains (\%)}
  \end{subfigure}
  \caption{Performance improvements over teacher-student baseline with different model architectures on NCI109.}
  \label{fig:arch_analysis_main}
\end{figure}

\subsubsection{Robustness to Noisy Explanations}
In real-world applications, providers may intentionally introduce noise into explanation graphs to hinder model extraction, or noise may inherently exist due to limitations of the deployed GNN explainer. The noises in explanation graphs could pose significant challenge to the model stealing with explanations. Hence, we simulate unreliable explanation scenarios by randomly shuffling a fraction of the target model's explanation outputs to disrupt the original node importance ranking. Figure~\ref{fig:noise_impact_sub} shows the AUC performance under different noise levels (0\%, 10\%, 20\%, 30\%) on NCI109. Our method demonstrates graceful degradation with increasing noise while maintaining advantages over the TS baseline across all noise levels. This demonstrates that our method maintains effectiveness even with noisy explanations. More results on other datasets are provided in Appendix~\ref{app:explanation_quality_results}.

\subsubsection{Adaptation to Binary Explanations}
\begin{figure}[t]
  \centering
  \begin{subfigure}[b]{0.49\linewidth}
    \centering
    \includegraphics[width=\linewidth]{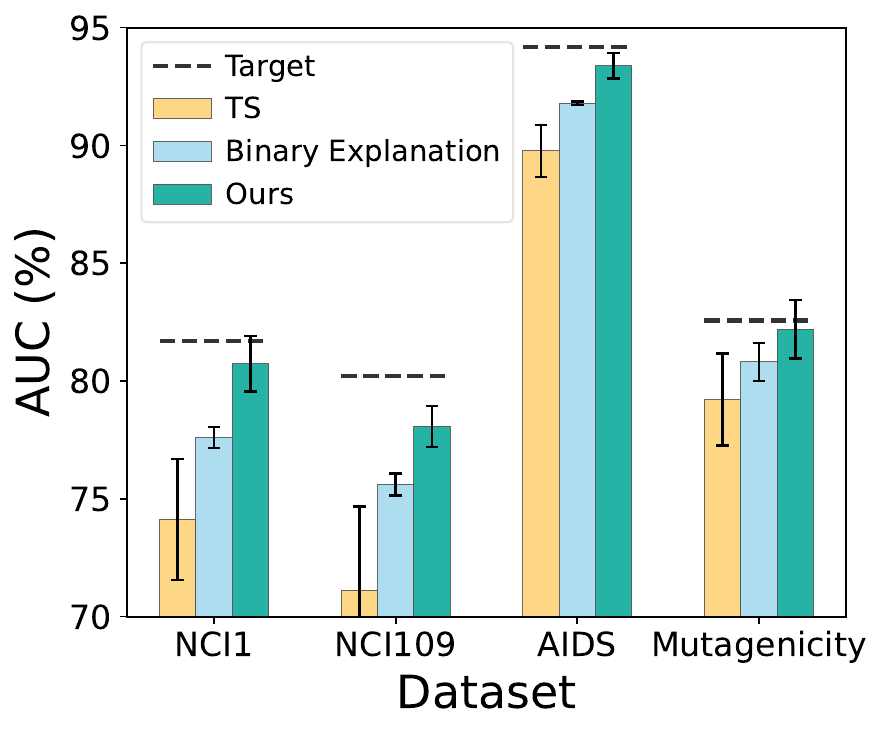}
    \caption{AUC}
  \end{subfigure}
  \hfill
  \begin{subfigure}[b]{0.49\linewidth}
    \centering
    \includegraphics[width=\linewidth]{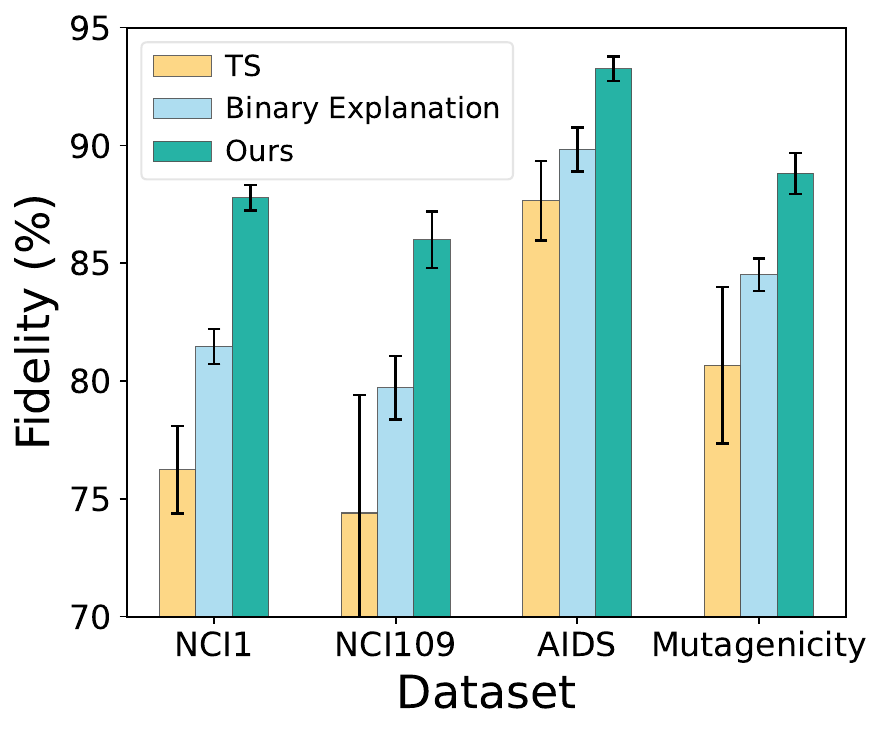}
    \caption{Fidelity}
  \end{subfigure}
  \caption{Model Stealing Results on Binary Explanations.}
  \label{fig:binary_explanation_sub}
\end{figure}

Some scenarios only provide coarse-grained explanations, such as binary indicators distinguishing important from irrelevant substructures. Figure~\ref{fig:binary_explanation_sub} shows that our method maintains advantages over TS baselines with this simplified format across all datasets. While performance naturally decreases compared to fine-grained explanations, the framework remains effective across different explanation granularities. Complete results for more metrics are given in Appendix~\ref{app:explanation_quality_results}.

\subsection{Flexibility of {\method}}

To answer the \textbf{RQ3}, we first conduct experiments on stealing GNN model for node classification. Then, we investigate the impacts of model architectures and explanation mechanisms to {\method}.

\subsubsection{Flexibility to Node Classification Tasks}
Beyond graph-level tasks, we evaluate our framework on node classification using PubMed and ogbn-arxiv datasets with the same data splitting strategy as in Section~\ref{subsection:experimental_setups}, with the query budget set to 10\% of the target model's training data size. To adapt node classification to our framework, we extract 2-hop neighborhood subgraphs centered on each node to transform the task into graph classification. Table~\ref{tab:node_classification_task} shows that our method outperforms baseline approaches across most evaluation metrics, demonstrating effective generalization across different graph learning tasks.

\subsubsection{Flexibility to Target/Surrogate Model Architectures} 

We investigate how different GNN architectures affect extraction performance by examining five architectures: GIN, GCN, GAT, GraphSAGE (GSAGE), and Graph Transformer (GT)~\cite{rampavsek2022recipe}. Figure~\ref{fig:arch_analysis_main} shows the AUC and fidelity gains of our method over the TS baseline on the NCI109 dataset. Performance improvements are observed regardless of whether the target and surrogate models share the same architecture, demonstrating the generalization capability of our explanation-guided approach across diverse architectural combinations. Complete results including explanation rank correlation are provided in Appendix~\ref{app:arch_analysis}.

\begin{table}[t]
    \centering
    \caption{Performance of different explanation mechanisms. Gray numbers show performance gains over the TS baseline.}
    \vspace{-0.5em}
    \label{tab:explainer_nci109}
    \resizebox{\linewidth}{!}{
      \begin{tabular}{lccc}
        \toprule
        \multirow{1}{*}{Explainer} & AUC (\%) & Fidelity (\%) & Rank Corr. (\%) \\
        \midrule
        CAM       & $78.08\pm0.87$ \textcolor{gray}{\scriptsize~↑6.95}   & $85.99\pm1.20$ \textcolor{gray}{\scriptsize~↑11.59} & $35.58\pm0.33$ \textcolor{gray}{\scriptsize~↑22.67} \\
        Grad      & $75.38\pm1.08$ \textcolor{gray}{\scriptsize~↑4.25}   & $78.50\pm0.94$ \textcolor{gray}{\scriptsize~↑4.10}  & $52.25\pm1.01$ \textcolor{gray}{\scriptsize~↑48.40} \\
        Grad-CAM  & $77.38\pm1.10$ \textcolor{gray}{\scriptsize~↑6.25}   & $85.43\pm2.30$ \textcolor{gray}{\scriptsize~↑11.03} & $36.60\pm0.79$ \textcolor{gray}{\scriptsize~↑23.68} \\
        GNNExpl.  & $75.12\pm1.59$ \textcolor{gray}{\scriptsize~↑3.99}   & $77.75\pm1.63$ \textcolor{gray}{\scriptsize~↑3.35}  & $20.25\pm0.62$ \textcolor{gray}{\scriptsize~↑22.76} \\
        PGExpl.   & $75.48\pm1.28$ \textcolor{gray}{\scriptsize~↑4.35}   & $77.99\pm2.31$ \textcolor{gray}{\scriptsize~↑3.59}  & $64.07\pm0.79$ \textcolor{gray}{\scriptsize~↑65.03} \\
        \bottomrule
      \end{tabular}
    }
\end{table}


\subsubsection{Flexibility to Explainers} 

We assess our framework's performance across five explanation methods applied to the target model: CAM, Grad, Grad-CAM, GNNExplainer, and PGExplainer. The experiments are conducted on NCI109 with the same setting as in Section~\ref{subsection:model_extraction_performance}. For PGExplainer, which generates edge importance scores, we convert them to node importance scores by averaging over adjacent edges for each node. As shown in Table~\ref{tab:explainer_nci109}, all explanation methods lead to performance improvements over the TS baseline across all three metrics, demonstrating that our approach generalizes well across different explanation techniques. Results for additional datasets are provided in Appendix~\ref{app:explainer_results}.

\subsection{Ablation Study}
In this subsection we conduct an ablation study by removing data augmentation and explanation alignment from our framework. Figure~\ref{fig:ablation_study_main} shows the AUC performance across four datasets, with complete results available in Appendix~\ref{app:ablation_results}. Our experiments demonstrate that both components enhance the framework's performance. Explanation alignment plays a crucial role in transferring decision logic, with its removal causing the most substantial performance drop. Data augmentation primarily improves predictive accuracy, leading to higher AUC and fidelity scores. When both components are removed, the framework essentially reverts to the TS baseline.

\begin{figure}[t]
  \centering
  \begin{subfigure}[b]{0.49\linewidth}
    \centering
    \includegraphics[width=\linewidth]{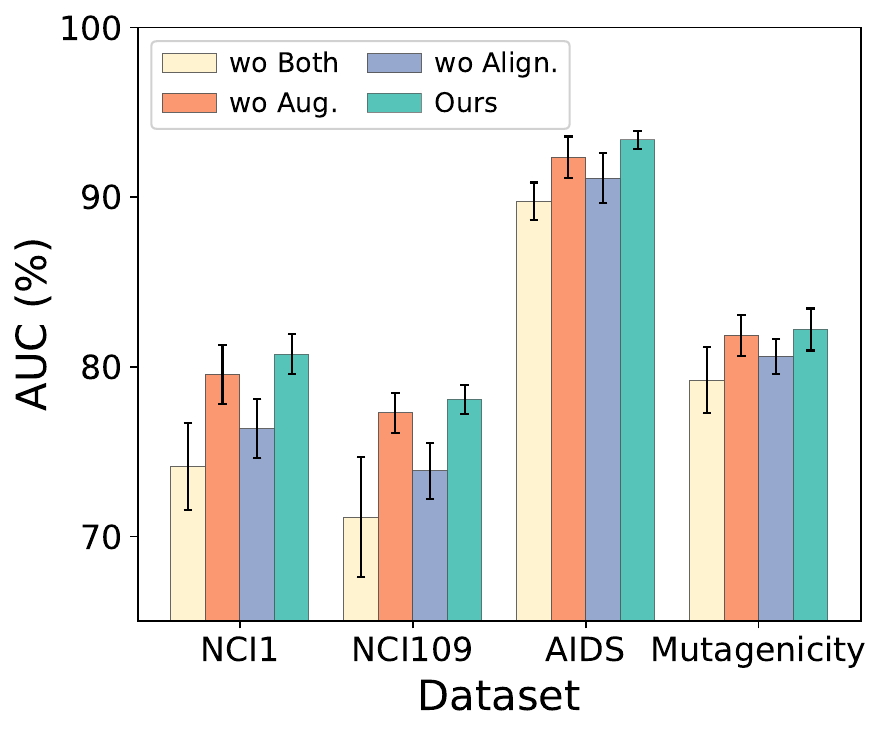}
    \caption{Impact on AUC}
  \end{subfigure}
  \hfill
  \begin{subfigure}[b]{0.49\linewidth}
    \centering
    \includegraphics[width=\linewidth]{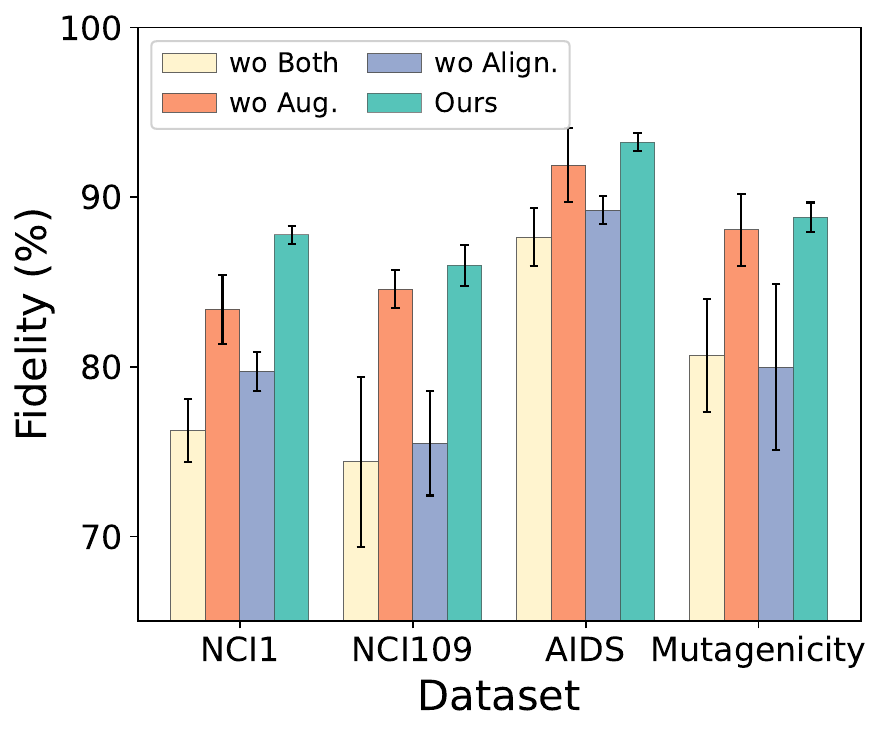}
    \caption{Impact on Fidelity}
  \end{subfigure}
  \caption{Ablation study results across four datasets showing the impact of removing data augmentation and explanation alignment components from our full framework.}
  \label{fig:ablation_study_main}
\end{figure}

\begin{figure}[t]
  \centering
  \begin{subfigure}[b]{0.49\linewidth}
    \centering
    \includegraphics[width=\linewidth]{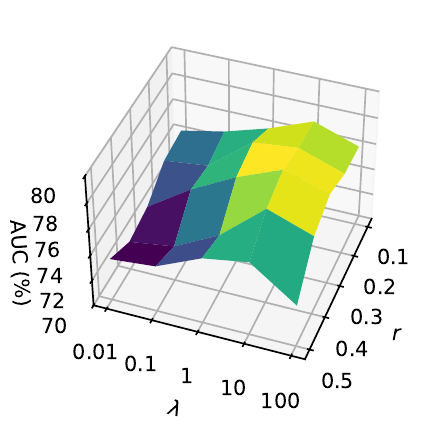}
    \caption{Impact on AUC}
  \end{subfigure}
  \hfill
  \begin{subfigure}[b]{0.49\linewidth}
    \centering
    \includegraphics[width=\linewidth]{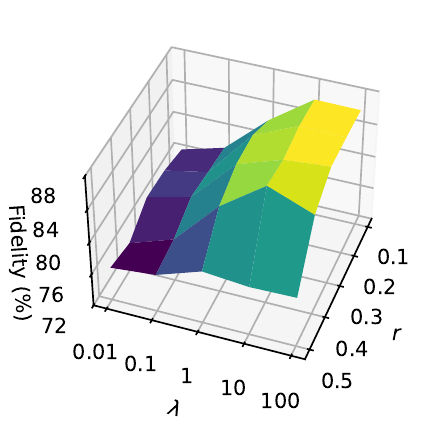}
    \caption{Impact on fidelity}
  \end{subfigure}
  \caption{Sensitivity analysis of data augmentation ratio $r$ and alignment loss coefficient $\lambda$.}
  \label{fig:param_analysis_main}
\end{figure}

\subsection{Hyperparameter Analysis}
\label{subsection:hyperparameter}
We analyze the sensitivity of two key hyperparameters: the data augmentation ratio $r$ and the alignment loss coefficient $\lambda$. We vary $r$ from 0.1 to 0.5 and $\lambda$ from 0.01 to 100, conducting experiments on the NCI109 dataset with settings consistent with Section~\ref{subsection:model_extraction_performance}. As shown in Figure~\ref{fig:param_analysis_main}, moderate augmentation ratios achieve the best AUC performance, suggesting that neither too little nor too much data augmentation is optimal. For the alignment coefficient $\lambda$, increasing it enhances fidelity and explanation alignment, but excessively large values may compromise predictive performance. Complete sensitivity analysis across all metrics is provided in Appendix~\ref{appendix:hyperparameter}.

\section{Conclusion}
\label{sec:conclusion}

This paper investigated the security vulnerabilities of explainable GNNs to model stealing attacks. We proposed {\method}, a novel stealing framework that exploits explanation information through two complementary components: a rank-based explanation alignment mechanism that effectively captures the target model's decision logic, and an explanation-guided data augmentation strategy that preserves essential causal patterns while enabling efficient training sample generation. Experiments across molecular datasets demonstrated that {\method} consistently outperforms existing approaches in capturing both predictive behavior and underlying decision logic. These findings reveal an important security-transparency trade-off in the deployment of explainable GNNs in high-stakes domains such as drug discovery. Future work will extend our investigation to more graph learning tasks and develop effective defense mechanisms against explanation-based model stealing attacks.

\section{Acknowledgment}

This material is based upon work supported by, or in part by, Prevention and Control of Emerging and Major Infectious Diseases-National Science and Technology Major Project (Grant No. 2026ZD\allowbreak01910200), the National Natural Science Foundation of China (NSFC) under Grant No. 62506316, and the Guangdong Provincial Program under Grant No. 2025D03J0015. The findings in this paper do not necessarily reflect the views of the funding agencies.

\bibliographystyle{ACM-Reference-Format}
\bibliography{Reference}

\appendix
\clearpage

\section{Details of Related Works}

\subsection{Explainable Graph Neural Networks}
\label{app:related_works1}
Graph Neural Networks (GNNs) have demonstrated remarkable success across various domains, including financial analysis~\cite{harl2020explainable, wang2021review, lv2019auto}, recommendation systems~\cite{chen2022grease, wu2022graph, hamilton2017inductive}, and biological analysis~\cite{li2021braingnn, sun2020graph, xiong2021graph, kawahara2017brainnetcnn}. The success of GNNs largely relies on their message-passing mechanism, where node representations are iteratively updated by aggregating information from their neighbors~\cite{xu2018powerful, kipf2016semi, velivckovic2017graph}. This enables GNNs to effectively capture both node features and graph structural information, leading to remarkable performance in various tasks including node classification, link prediction, and graph classification.

To enhance the interpretability of these sophisticated models, researchers have proposed various explanation methods~\cite{yuan2022explainability,dai2021towards, dai2025towards}. Early approaches follow methods in computer vision to use gradient and attribution-based techniques, such as sensitivity analysis and guided backpropagation~\cite{baldassarre2019explainability} utilize gradient information to identify important input features, while Class Activation Mapping (CAM)~\cite{zhou2016learning, pope2019explainability} generates explanations by examining the final convolutional layer's feature maps. Grad-CAM~\cite{selvaraju2017grad, pope2019explainability} further improves upon CAM by incorporating gradient information to better localize important regions in the input graph. These methods, though intuitive, often struggle to capture complex structural dependencies in GNNs.

To address these limitations, more sophisticated approaches have been developed. GNNExplainer~\cite{ying2019gnnexplainer} formulates explanation generation as an optimization problem that identifies a compact subgraph maximizing the mutual information with the model's prediction. PGExplainer~\cite{luo2020parameterized} extends this idea by training a parameterized explanation generator that learns to identify important edges across multiple instances. GraphLIME~\cite{huang2022graphlime} adapts the concept of local interpretable model-agnostic explanations to graph data, while SubgraphX~\cite{yuan2021explainability} employs monte carlo tree search and Shapley values to generate hierarchical explanations. GraphMask~\cite{schlichtkrull2020interpreting} takes a different approach by learning to identify dispensable edges during message passing. These methods have demonstrated increasing effectiveness in providing insights into GNN predictions, though their deployment raises important questions about the potential leakage of model decision logic compared to traditional black-box predictions.

\subsection{Model Stealing Attacks against GNNs}
\label{app:related_works2}
As an important aspect of privacy attacks, model stealing or model extraction attacks aim to extract the target model information by learning
a surrogate model that behaves similarly to the target model \cite{dai2022comprehensive, tramer2016stealing}. Generally, the attacker will first query the APIs of the target model to obtain predictions on the shadow dataset. It then leverages the shadow dataset and the corresponding predictions to train the surrogate model for model extraction attack \cite{tramer2016stealing}.
Following \cite{tramer2016stealing}, recent works have developed various model extraction attacks against graph neural networks. Wu et al.~\cite{wu2022model} proposed the first framework of GNN model extraction attacks along multiple dimensions based on the attacker's different background knowledge of the shadow dataset: access to node attributes, knowledge of graph structure, and availability of shadow graphs. This taxonomy framework helps understand different attack scenarios in real-world applications, though their focus remains on transductive settings where attackers can access training processes.
Shen et al.~\cite{shen2022model} made a significant advance by proposing the first model-stealing attacks against inductive GNNs, which better reflects real-world deployment scenarios. They increased the amount of information extracted from the victim model by aligning not only the predictions but also other responses such as node embeddings or prediction logits (soft label).  Building upon this direction, Podhajski et al. \cite{podhajski2024efficient} further augmented the node embeddings with graph contrastive learning and spectral graph augmentations, and Zhuang et al.~\cite{zhuang2024unveiling} recently proposed STEALGNN, introducing a more challenging scenario where attackers have no access to any real graph data. Their work demonstrates the possibility of model extraction through carefully designed synthetic graph generation.

With the popularity of explainable AI, early research in other fields has demonstrated that explanations can reveal information about model parameters~\cite{aivodji2020model, miura2024megex, yan2023explanationdata}.  Milli et al.~\cite{milli2019model} demonstrated that gradient-based explanations of a model can reveal the model itself. Their theoretical analysis proved that with gradient information, the number of queries needed to reconstruct a model can be significantly reduced compared to prediction-only approaches. Yan et al. \cite{yan2023explanation} further introduced an extra CNN autoencoder to utilize the representations learned by reconstructing the explanations. Other studies explore the possibility of data-free model extraction with explainable AI~\cite{yan2023explanationdata}. However, the unique graph structures and message-passing mechanisms in GNNs significantly differentiate their decision-making and explanation processes. Thus, previous methods for images are not directly applicable to explainable GNNs. With the widespread adoption of explainable GNNs in critical areas such as drug discovery, with explanation methods like GNNExplainer \cite{ying2019gnnexplainer}, it is necessary to address the \emph{security risks of model extraction attacks arising from explanations}. To the best of our knowledge, we are the \emph{first} to investigate the susceptibility of GNNs to model extraction attacks caused by model explanations.

\section{Details of Explanation Approaches}
\label{appendix:explanation_methods}

We provide the detailed formulations of the explanation methods used in our experiments. Let $\mathbf{F}^{(L)} \in \mathbb{R}^{|\mathcal{V}| \times d}$ denote the node representations from the final layer $L$ of the GNN encoder, where $d$ is the hidden dimension. For any node $v \in \mathcal{V}$, we use $\mathbf{F}^{(L)}_{k,v}$ to denote its $k$-th feature.

\paragraph{CAM} The Graph Class Activation Mapping generates node importance scores in three steps. First, it performs global average pooling on the final layer representations:
\begin{equation}
    e_k = \frac{1}{|\mathcal{V}|}\sum_{v \in \mathcal{V}} \mathbf{F}^{(L)}_{k,v},
\end{equation}
Then, the class score for class $c$ is computed as:
\begin{equation}
    y_c = \sum_k w^c_k e_k,
\end{equation}
where $w^c_k$ represents the weight connecting the $k$-th feature to class $c$ in the classification layer. Finally, the importance score for node $v$ is obtained by:
\begin{equation}
    \mathbf{E}_v = \sum_k w^c_k \mathbf{F}^{(L)}_{k,v}.
\end{equation}

\paragraph{Gradient-based} This method computes node importance scores by taking the gradients of the target class score $y_c$ with respect to the node features in the final layer:
\begin{equation}
    \mathbf{E}_v = \|\text{ReLU}(\frac{\partial y_c}{\partial \mathbf{F}^{(L)}_v})\|,
\end{equation}
where $\mathbf{F}^{(L)}_v$ denotes the feature vector of node $v$ at layer $L$.

\paragraph{Grad-CAM} The Gradient-weighted Class Activation Mapping first computes class-specific weights by averaging the gradients:
\begin{equation}
    \alpha^c_k = \frac{1}{|\mathcal{V}|}\sum_{v \in \mathcal{V}} \frac{\partial y_c}{\partial \mathbf{F}^{(L)}_{k,v}},
\end{equation}
The node importance scores are then calculated as:
\begin{equation}
    \mathbf{E}_v = \text{ReLU}(\sum_k \alpha^c_k \mathbf{F}^{(L)}_{k,v}).
\end{equation}

\paragraph{GNNExplainer and PGExplainer} For these methods, we use their official implementations. Since PGExplainer generates edge importance scores $\mathbf{E}_{(u,v)}$, we convert them to node importance scores by averaging over the adjacent edges:
\begin{equation}
    \mathbf{E}_v = \frac{1}{|\mathcal{N}(v)|}\sum_{u \in \mathcal{N}(v)} \mathbf{E}_{(u,v)},
\end{equation}
where $\mathcal{N}(v)$ denotes the neighbors of node $v$.

\section{Computational Complexity}
\label{app:complexity}

We analyze the computational complexity of our explanation-guided model stealing framework. Let $Q$ denote the query budget, $|\mathcal{V}|$ and $|\mathcal{E}|$ represent the average number of nodes and edges in each graph, $L$ be the number of GNN layers, $d$ be the hidden dimension of neural networks, and $r$ be the data augmentation ratio.

For the data collection phase, we perform $Q$ queries to the target model, obtaining predictions and explanations with a cost of $O(Q \cdot |\mathcal{V}|)$ for processing the explanations. The subsequent data augmentation generates $r \cdot Q$ additional training samples without requiring further queries. Since each augmentation operation (node dropping, edge perturbation, or subgraph extraction) needs to process the graph structure, it requires $O(|\mathcal{V}| + |\mathcal{E}|)$ operations per graph, resulting in a total augmentation complexity of $O(r \cdot Q \cdot (|\mathcal{V}| + |\mathcal{E}|))$.

During the model training phase, we process a total of $(1+r)Q$ samples over $T$ iterations. For each sample, the GNN forward and backward propagation requires $O(L \cdot d \cdot (|\mathcal{V}| + |\mathcal{E}|))$ operations, where the term $(|\mathcal{V}| + |\mathcal{E}|)$ accounts for message passing through all nodes and edges. The loss computation involves the cross-entropy prediction loss with $O(c)$ complexity and the ranking-based explanation alignment loss with $O(|\mathcal{V}|^2)$ complexity due to pairwise comparisons of node importance. Considering all $(1+r)Q$ samples and $T$ iterations, the overall training complexity becomes $O(T \cdot (1+r)Q \cdot (L \cdot d \cdot (|\mathcal{V}| + |\mathcal{E}|) + c + |\mathcal{V}|^2))$.

The dominant computational factor in our approach is the quadratic term $O(|\mathcal{V}|^2)$ from the explanation alignment loss, which arises from pairwise comparisons of node importance rankings. While this introduces additional cost compared to standard GNN training, it is essential for effectively capturing the target model's reasoning process, and the overhead remains manageable in the considered task settings. For graph classification, $|\mathcal{V}|$ corresponds to the size of each input graph, which is typically small in molecular graph datasets. For node classification on large graphs, since predictions and explanations are computed on local $k$-hop neighborhoods around queried nodes, $|\mathcal{V}|$ denotes the size of the effective local subgraph rather than the entire graph, thereby mitigating the scalability concern.

\section{Experimental Setting}
\label{app:experimental_setting}

\subsection{Evaluation Metrics Details}
\label{appendix:metrics}

We employ three metrics to evaluate the effectiveness of our model stealing framework:

\begin{itemize}[leftmargin=*]
    \item \textbf{Area Under the ROC Curve (AUC)} on the test set, which measures the predictive performance of the surrogate model.
    
    \item \textbf{Prediction Fidelity}, which measures the agreement between the predictions of the surrogate and target models:
    \begin{equation}
        \text{Fidelity} = \frac{1}{|\mathcal{D}|} \sum_{\mathcal{G} \in \mathcal{D}} \mathds{1} [y_{\theta'}(\mathcal{G}) = y_{\theta}(\mathcal{G})],
    \end{equation}
    where $y_{\theta'}(\mathcal{G})$ and $y_{\theta}(\mathcal{G})$ denote the predictions from the surrogate and target models respectively, and $\mathcal{D}$ is the test dataset.

    \item \textbf{Rank Correlation}, which quantifies the alignment of decision logic between the surrogate and target models through the Kendall's tau coefficient:
    \begin{equation}
        \tau = \frac{1}{|\mathcal{D}|} \sum_{\mathcal{G} \in \mathcal{D}} \text{corr}_{\tau}(\mathbf{E}_{\theta'}(\mathcal{G}), \mathbf{E}_{\theta}(\mathcal{G})),
    \end{equation}
    where $\text{corr}_{\tau}$ denotes the Kendall's tau coefficient, and $\mathbf{E}_{\theta'}(\mathcal{G})$ and $\mathbf{E}_{\theta}(\mathcal{G})$ are explanations from the surrogate and target models for graph $\mathcal{G}$, respectively.
\end{itemize}

\subsection{Model Hyperparameters and Training Details}
\label{appendix:hyperparameters_and_training_details}
Our target models implemented various architectures including GIN, GCN, GAT, GraphSAGE, and Graph Transformer, while surrogate models consistently used GIN architecture. All models were configured with 3 GNN layers and 128 hidden dimensions, trained using Adam optimizer (learning rate 0.001) with batch size 64. Target models were trained for 200 epochs with checkpoints selected based on validation AUC. As for pretrained target models, we used a 5-layer GIN with 300-dimensional embeddings, fine-tuned for 100 epochs. Our data splitting strategy allocated 40\% for target model training or fine-tuning (further split into 80\% train, 20\% validation), 40\% for shadow data, and 20\% for test data. All experiments were conducted on a single NVIDIA RTX A6000 GPU and repeated five times with different random seeds [41, 42, 43, 44, 45], with final results reported as means with standard deviations.

\subsection{Compared Methods} 
\label{appendix:compared_methods}

Firstly, we include the fundamental teacher-student framework of the model extraction attack:

\begin{itemize}[leftmargin=*]
    \item \textbf{Teacher-Student} \cite{tramer2016stealing} is the fundamental model extraction baseline to train a surrogate model via the target model's input data and prediction APIs.
\end{itemize}

We include recent works on model stealing attacks targeting GNNs:

\begin{itemize}[leftmargin=*]
\item  \textbf{MEA-GNN} \cite{wu2022model} provides a taxonomy of GNN extraction attacks on transductive node classification tasks, categorized into 7 attacks based on the available knowledge about the target dataset and query graphs. We adapt this framework's Attack-3 to our  graph classification setting, where query graphs are known but other training data remains unknown.  

\item \textbf{GNNStealing} \cite{shen2022model} introduces two types of attacks where the attack can obtain the query graphs (type I) and when the graph structural information
is missing (type II). They also include the predicted logits (soft labels) in complement to the hard labels to enhance the performance. Since we can access the complete query graph in our setting, we adapt their type I attack and use soft logits as responses for our graph classification setting.

\item \textbf{EfficientGNN} \cite{podhajski2024efficient} followed GNNStealing and further enriched the learning of node
embeddings with graph contrastive learning. Additionally, they introduced a spectral graph
augmentation method to increase the stealing efficiency. We adapt their data augmentation strategies for comparison.

\item \textbf{STEALGNN} \cite{zhuang2024unveiling} introduces a graph generator to generate query graphs for data-free GNN extraction attacks. During training, the graph generator and surrogate model are trained adversarially to generate diverse graphs that better capture the victim model's decision-making process. We adapt their approach to our setting by combining both the generated graphs and our query graphs to train the surrogate model.

\end{itemize}

We also include recent studies on leveraging explanations for model extraction in other domains to offer a broader perspective:

\begin{itemize}[leftmargin=*]
\item \textbf{MRME} \cite{milli2019model} explored the possibility of quicker model retrieval through gradient-based explanations. We adapt this method to our GNN setting by directly aligning the gradient-based explanations of graph neural networks between the target and surrogate model.

\item \textbf{DET} \cite{yan2023explanation} introduced an autoencoder to reconstruct the original image and an extra autoencoder reconstruct their explanations. By combining the embeddings of two encoders, they use the features learned from explanations to train a surrogate classifier.
\end{itemize}

\section{Implementation Details of Cross-Distribution Experiments}
\label{appendix:cross_distribution_implementation}

\subsection{Feature Dimension Alignment}
\label{appendix:feature_align}

In our cross-distribution experiments, we encounter feature dimension misalignment between different molecular graph datasets. Specifically, while NCI109 and AIDS datasets have 38-dimensional node features, NCI1 has 37-dimensional features. To enable cross-dataset model stealing attacks while preserving the original feature information, we pad the NCI1 features with an additional zero dimension:
\begin{equation}
    \mathbf{x}_{NCI1}^{new} = [\mathbf{x}_{NCI1}; 0] \in \mathbb{R}^{38},
\end{equation}
where $\mathbf{x}_{NCI1} \in \mathbb{R}^{37}$ is the original feature vector and $[\cdot;\cdot]$ denotes feature concatenation. This padding approach maintains the intrinsic feature patterns of NCI1 while ensuring compatible input dimensions for the GNN models.

\subsection{Maximum Mean Discrepancy Computation}
\label{appendix:mmd}

To quantify the distribution differences between datasets, we compute the Maximum Mean Discrepancy (MMD) using a 16-dimensional feature vector that captures key structural properties of each graph. The MMD is defined as:
\begin{equation}
\begin{split}
    \text{MMD}^2(\mathcal{P}, \mathcal{Q}) = & \mathbb{E}_{x,x' \sim \mathcal{P}}[k(x,x')] + \mathbb{E}_{y,y' \sim \mathcal{Q}}[k(y,y')] \\
    & - 2\mathbb{E}_{x \sim \mathcal{P}, y \sim \mathcal{Q}}[k(x,y)],
\end{split}
\end{equation}
where $\mathcal{P}$ and $\mathcal{Q}$ are two distributions, and $k(\cdot,\cdot)$ is a kernel function.

For each graph, we extract the following features:

1) \textbf{Degree Distribution Statistics} (5 features):
   - Mean node degree
   - Standard deviation of node degrees
   - 25th percentile of node degrees
   - Median node degree
   - 75th percentile of node degrees

2) \textbf{Clustering Coefficient Statistics} (5 features):
   - Mean clustering coefficient
   - Standard deviation of clustering coefficients
   - 25th percentile of clustering coefficients
   - Median clustering coefficient
   - 75th percentile of clustering coefficients

3) \textbf{Graph Diameter} (1 feature):
   - Diameter of the largest connected component

4) \textbf{Spectral Features} (5 features):
   - Top 5 eigenvalues of the adjacency matrix
   - For graphs with fewer than 5 nodes, we pad with zeros to maintain fixed dimensionality

After extracting these 16-dimensional feature vectors for all graphs in both datasets, we normalize them using z-score normalization:
\begin{equation}
    \mathbf{x}_{normalized} = \frac{\mathbf{x} - \mu}{\sigma + \epsilon},
\end{equation}
where $\mu$ and $\sigma$ are the mean and standard deviation computed across all graphs from both datasets, and $\epsilon=10^{-8}$ is added for numerical stability.

We then compute the MMD using a Gaussian kernel:
\begin{equation}
    k(x,y) = \exp(-\gamma ||x-y||^2),
\end{equation}
where $\gamma=0.5$ is the kernel bandwidth parameter. Using this approach, we compute the MMD between NCI109 and NCI1 datasets, as well as between NCI109 and AIDS datasets, to quantify their distribution differences.

\section{Additional Results}

\subsection{Additional Results with Different Target Model Architectures}
\label{app:additional_results_target_models}
To demonstrate the effectiveness of our approach across different target model architectures, we conduct additional experiments with GCN and GAT as target models. All surrogate models use the GIN architecture, and we maintain the same experimental setup as described in Section~\ref{subsection:model_extraction_performance}.

\begin{table*}[ht]
  \caption{Performance comparison with GCN target models trained from scratch}
  \vspace{-0.5em}
  \centering
  \resizebox{\linewidth}{!}{
  \begin{tabular}{llccccccccc}
    \toprule
    Dataset & Metric (\%) & Target & TS & MEA-GNN & GNNStealing & EfficientGNN & MRME & DET & STEALGNN & Ours \\
    \midrule
    \multirow{3}{*}{NCI1}
    & AUC         & $74.68$ & $73.17 \pm 1.33$ & $71.89 \pm 0.83$ & $71.87 \pm 1.03$ & $72.77 \pm 1.00$ & \underline{$73.23 \pm 1.27$} & $70.88 \pm 0.60$ & $71.36 \pm 0.69$ & \textbf{73.64 $\pm$ 0.49} \\
    & Fidelity    &  \multicolumn{1}{c}{--}    & $85.35 \pm 2.17$ & $84.43 \pm 1.73$ & $83.87 \pm 1.51$ & $84.94 \pm 1.50$ & \underline{$85.89 \pm 2.29$} & $80.44 \pm 0.90$ & $85.56 \pm 0.72$ & \textbf{85.89 $\pm$ 1.54} \\
    & Rank Corr.  & \multicolumn{1}{c}{--}   & $10.68 \pm 1.34$ & $9.92 \pm 3.29$  & $10.40 \pm 2.63$ & $9.42 \pm 2.18$  & \underline{$10.77 \pm 1.85$} & $6.17 \pm 3.69$ & $9.46 \pm 1.65$ & \textbf{37.00 $\pm$ 0.60} \\
    \midrule
    \multirow{3}{*}{NCI109}
    & AUC         & $72.66$ & \underline{$71.13 \pm 1.08$} & $70.76 \pm 0.68$ & $70.72 \pm 1.24$ & $70.17 \pm 0.86$ & $71.11 \pm 1.08$ & $68.91 \pm 1.65$ & $69.56 \pm 1.03$ & \textbf{72.22 $\pm$ 0.34} \\
    & Fidelity    & \multicolumn{1}{c}{--}    & $86.88 \pm 0.99$ & \textbf{88.41 $\pm$ 0.98} & \underline{$88.02 \pm 3.32$} & $86.40 \pm 1.88$ & $87.05 \pm 0.84$ & $80.82 \pm 2.11$ & $87.47 \pm 0.37$ & $87.90\pm0.84$ \\
    & Rank Corr.  & \multicolumn{1}{c}{--}    & $13.58 \pm 0.97$ & \underline{$16.75 \pm 1.23$} & $8.97 \pm 2.62$ & $15.41 \pm 2.06$ & $13.73 \pm 0.73$ & $11.22 \pm 1.34$ & $13.86 \pm 0.91$ & \textbf{31.61 $\pm$ 0.28} \\
    \midrule
    \multirow{3}{*}{AIDS}
    & AUC         & $87.96$ & $78.86 \pm 4.19$ & $80.04 \pm 2.09$ & $80.24 \pm 4.60$ & $78.44 \pm 4.03$ & $79.01 \pm 4.29$  & \textbf{85.49 $\pm$ 2.34} & $83.02 \pm 0.66$ & \underline{$84.53\pm2.11$} \\
    & Fidelity   & \multicolumn{1}{c}{--}    & $91.60 \pm 3.17$ & $92.20 \pm 1.62$ & $90.40 \pm 0.93$ & \textbf{92.95 $\pm$ 0.53} & \underline{$92.65 \pm 1.31$} & $83.95 \pm 1.71$ & $89.99 \pm 15.41$ & $92.20\pm0.97$ \\
    & Rank Corr.  & \multicolumn{1}{c}{--}    & $11.15 \pm 4.56$ & $19.17 \pm 4.14$ & $12.57 \pm 6.97$ & $12.60 \pm 3.73$ & $11.38 \pm 4.12$ & $13.31 \pm 6.07$ & \underline{$26.82 \pm 3.48$} & \textbf{53.82 $\pm$ 0.93} \\
    \midrule
    \multirow{3}{*}{Mutagenicity}
    & AUC         & $84.00$ & $80.01 \pm 1.59$ & $81.17 \pm 0.76$ & \textbf{82.91 $\pm$ 0.56} & $79.69 \pm 1.28$ & $80.04 \pm 1.54$ & $79.42 \pm 0.97$ & $81.19 \pm 0.25$ & \underline{$81.58\pm0.79$} \\
    & Fidelity    & \multicolumn{1}{c}{--}  & $90.47 \pm 1.32$ & $89.57 \pm 2.40$ & \textbf{93.66 $\pm$ 1.22} & $89.99 \pm 1.71$ & $90.73 \pm 1.22$  & $84.57 \pm 0.69$ & $84.25 \pm 5.24$ & \underline{$92.11 \pm 0.79$} \\
    & Rank Corr.  & \multicolumn{1}{c}{--}   & $26.18 \pm 1.44$ & $24.97 \pm 2.85$ & $17.04 \pm 3.85$ & $25.70 \pm 1.86$ & $26.23 \pm 1.39$ & $21.11 \pm 2.93$ & \underline{$29.19\pm 0.23$} & \textbf{45.16 $\pm$ 0.98} \\
    \bottomrule
  \end{tabular}
  }
\end{table*}

\begin{table*}[ht]
  \caption{Performance comparison with GAT target models trained from scratch}
  \vspace{-0.5em}
  \centering
  \resizebox{\linewidth}{!}{
  \begin{tabular}{llccccccccc}
    \toprule
    Dataset & Metric (\%) & Target & TS & MEA-GNN & GNNStealing & EfficientGNN & MRME & DET & STEALGNN & Ours \\
    \midrule
    \multirow{3}{*}{NCI1}
    & AUC         & 79.00 & $73.62 \pm 1.00$ & $72.29 \pm 1.01$ & \underline{$74.68 \pm 0.58$} & $74.13 \pm 1.06$ & $73.91 \pm 0.95$ & $66.59 \pm 2.26$ & $70.55 \pm 1.72$ & \textbf{76.01 $\pm$ 0.44} \\
    & Fidelity    & --    & $82.77 \pm 1.10$ & $78.71 \pm 1.45$ & $84.48 \pm 1.52$ & \underline{$84.53 \pm 0.24$} & $82.87 \pm 1.31$ & $74.38 \pm 2.81$ & $79.76 \pm 1.39$ & \textbf{88.37 $\pm$ 0.84} \\
    & Rank Corr.  & --    & $14.15 \pm 1.04$ & $11.38 \pm 1.05$ & $10.65 \pm 2.52$ & \underline{$15.26 \pm 0.90$} & $14.34 \pm 0.99$ & $10.27 \pm 4.16$ & $12.49 \pm 2.09$ & \textbf{37.76 $\pm$ 0.52} \\
    \midrule
    \multirow{3}{*}{NCI109}
    & AUC         & 73.99 & $72.93 \pm 1.26$ & \underline{$73.24 \pm 0.68$} & $70.23 \pm 4.76$ & $73.03 \pm 0.49$ & $73.21 \pm 1.23$ & $70.08 \pm 1.03$ & $71.83 \pm 0.80$ & \textbf{74.56 $\pm$ 1.00} \\
    & Fidelity    & --    & $80.17 \pm 2.76$ & $80.00 \pm 4.32$ & $78.91 \pm 9.86$ & \underline{$83.27 \pm 1.65$} & $82.16 \pm 1.61$ & $75.76 \pm 1.24$ & $80.12 \pm 1.40$ & \textbf{85.99 $\pm$ 1.12} \\
    & Rank Corr.  & --    & $10.22 \pm 3.72$ & \underline{$13.35 \pm 2.44$} & $4.32 \pm 3.90$ & $11.97 \pm 3.13$ & $10.92 \pm 2.85$ & $10.02 \pm 3.20$ & $12.37 \pm 0.67$ & \textbf{30.50 $\pm$ 0.40} \\
    \midrule
    \multirow{3}{*}{AIDS}
    & AUC         & 89.89 & $88.05 \pm 0.98$ & $85.51 \pm 1.73$ & $86.69 \pm 2.49$ & \underline{$88.23 \pm 1.81$} & $88.15 \pm 1.14$ & $85.49 \pm 2.34$ & $85.12 \pm 0.17$ & \textbf{88.77 $\pm$ 0.61} \\
    & Fidelity    & --    & $86.40 \pm 1.33$ & $85.40 \pm 2.59$ & $85.50 \pm 2.49$ & \underline{$86.90 \pm 1.11$} & $86.20 \pm 1.56$ & $83.95 \pm 1.71$ & $82.91 \pm 0.51$ & \textbf{88.05 $\pm$ 0.70} \\
    & Rank Corr.  & --    & $13.41 \pm 3.72$ & \underline{$20.11 \pm 2.66$} & $15.54 \pm 1.55$ & $13.77 \pm 1.85$ & $13.40 \pm 3.69$ & $13.31 \pm 6.07$ & $14.52 \pm 3.30$ & \textbf{46.52 $\pm$ 1.35} \\
    \midrule
    \multirow{3}{*}{Mutagenicity}
    & AUC         & 84.06 & $80.81 \pm 0.97$ & $79.85 \pm 1.15$ & \underline{$82.53 \pm 0.79$} & $81.31 \pm 1.38$ & $80.83 \pm 1.03$ & $77.25 \pm 1.01$ & $81.05 \pm 0.25$ & \textbf{83.16 $\pm$ 0.71} \\
    & Fidelity    & --    & $85.40 \pm 1.05$ & $84.34 \pm 1.65$ & \textbf{89.57 $\pm$ 1.43} & $85.54 \pm 1.35$ & $85.35 \pm 1.16$ & $76.61 \pm 0.99$ & $84.74 \pm 0.68$ & \underline{$89.04 \pm 1.49$}\\
    & Rank Corr.  & --    & $22.54 \pm 1.77$ & $24.56 \pm 1.72$ & $19.17 \pm 7.48$ & $22.65 \pm 1.33$ & $22.32 \pm 1.57$ & $18.20 \pm 3.24$ & \underline{$24.60 \pm 1.46$} & \textbf{50.18 $\pm$ 0.49} \\
    \bottomrule
  \end{tabular}
  }
\end{table*}

\subsection{Additional Results on Varying Query Budgets}
\label{app:query_budget_analysis}

\begin{figure*}[ht]
  \centering
  \begin{subfigure}[b]{0.36\linewidth}
    \centering
    \includegraphics[width=\linewidth]{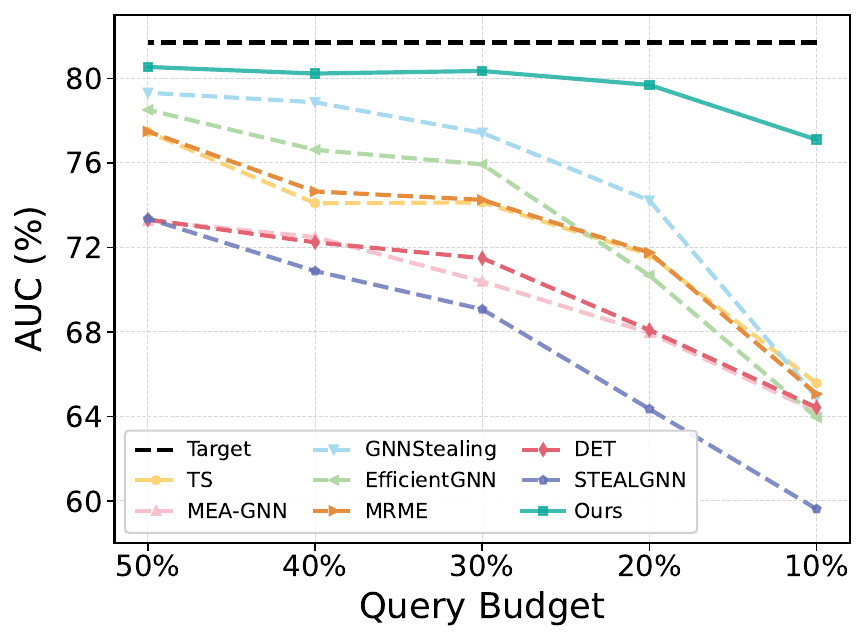}
    \caption{NCI1}
    \label{fig:full_budget_nci1}
  \end{subfigure}
  \qquad
  \begin{subfigure}[b]{0.36\linewidth}
    \centering
    \includegraphics[width=\linewidth]{figs/query_budget_NCI109.pdf}
    \caption{NCI109}
    \label{fig:full_budget_nci109}
  \end{subfigure}
  \qquad
  \begin{subfigure}[b]{0.36\linewidth}
    \centering
    \includegraphics[width=\linewidth]{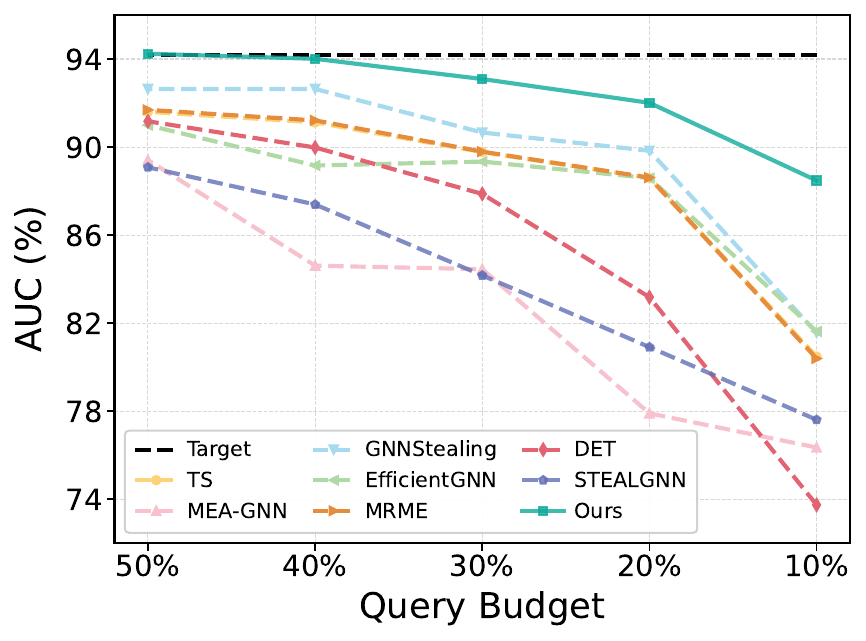}
    \caption{AIDS}
    \label{fig:full_budget_aids}
  \end{subfigure}
  \qquad
  \begin{subfigure}[b]{0.36\linewidth}
    \centering
    \includegraphics[width=\linewidth]{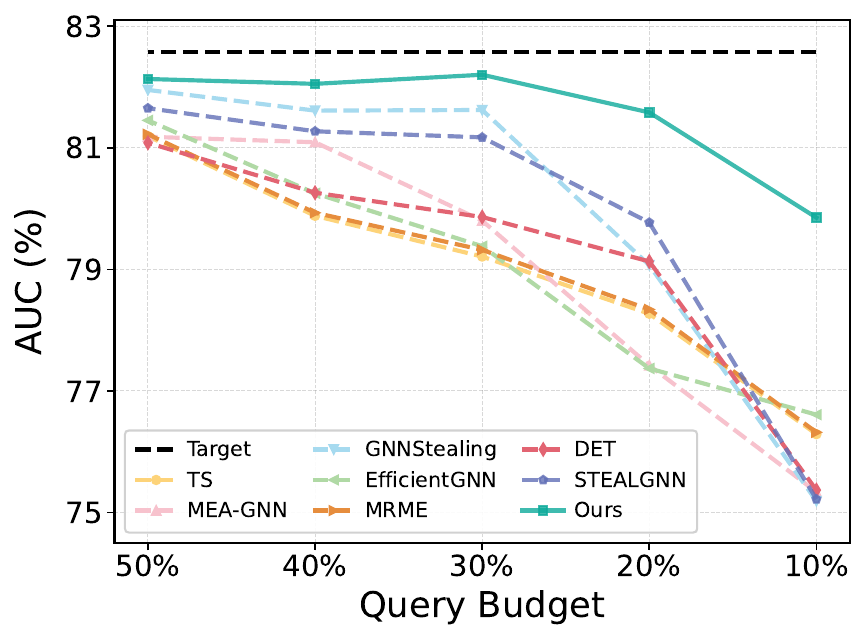}
    \caption{Mutagenicity}
    \label{fig:full_budget_mut}
  \end{subfigure}
  \vspace{-0.5em}
  \caption{Impact of query budget on extraction performance across datasets. The x-axis shows the percentage of target model's training data used for querying.}
  \label{fig:full_query_budget}
\end{figure*}

Figure~\ref{fig:full_query_budget} shows the extraction performance across NCI1, NCI109, AIDS, and Mutagenicity datasets when reducing query budgets from 50\% to 10\% of the target model's training data size. The results show that our approach exhibits more gradual performance degradation compared to baselines as query budgets decrease. Across all datasets, the performance gap between our method and baselines widens at lower query ratios, indicating that explanation-guided stealing makes more efficient use of limited queries. For instance, on NCI109, when the query budget drops from 50\% to 10\%, our method's AUC decreases by 4.11\% (from 78.64\% to 74.53\%), while the TS baseline drops by 9.23\% (from 74.55\% to 65.32\%). Similar patterns are observed on other datasets, where our method maintains relatively stable performance even under severely constrained query budgets. These results suggest that leveraging explanations helps extract more informative signals from each query, which becomes particularly beneficial when query opportunities are limited.

\subsection{Additional Results on Cross-Distribution Analysis}
\label{appendix:cross_distribution_results}
\begin{figure*}[ht]
  \centering
  \begin{subfigure}[b]{0.33\linewidth}
    \centering
    \includegraphics[width=\linewidth]{figs/cross_dist_AUC.pdf}
    \caption{AUC}
    \label{fig:full_cross_dist_AUC}
  \end{subfigure}
  \hfill
  \begin{subfigure}[b]{0.33\linewidth}
    \centering
    \includegraphics[width=\linewidth]{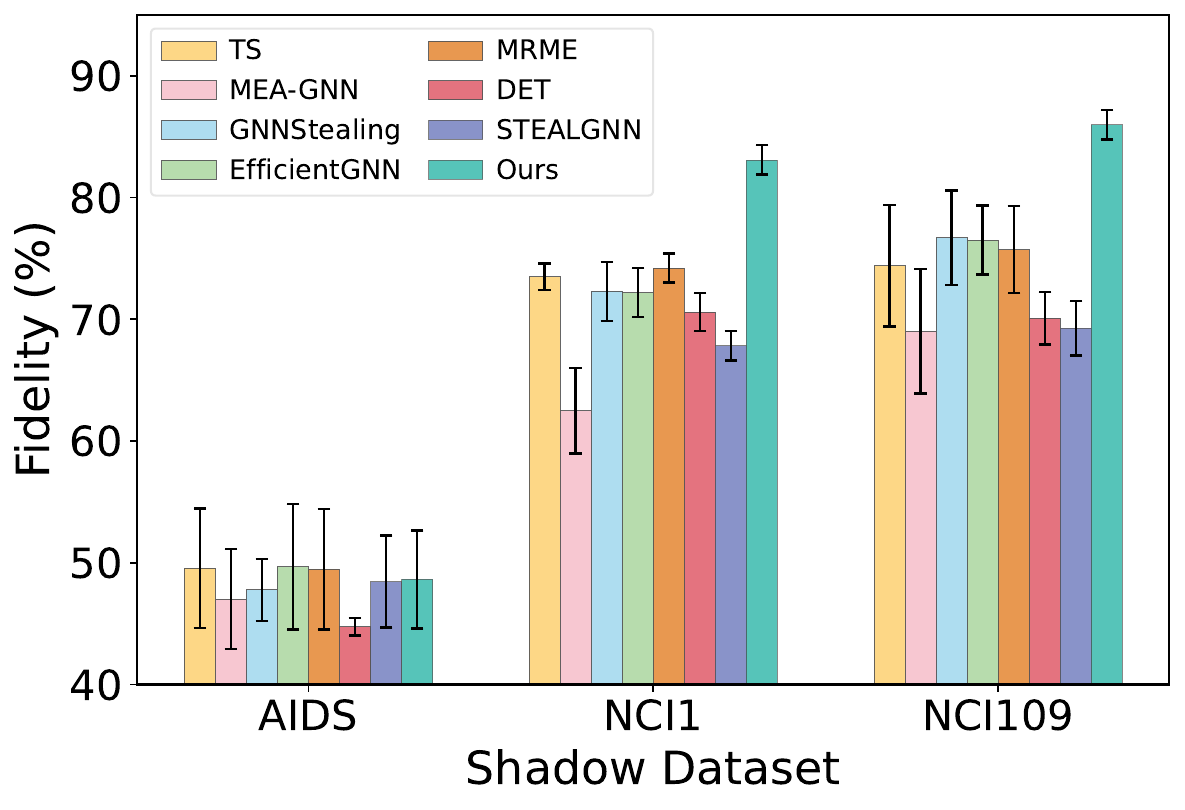}
    \caption{Fidelity}
    \label{fig:full_cross_dist_Fidelity}
  \end{subfigure}
  \hfill
  \begin{subfigure}[b]{0.33\linewidth}
    \centering
    \includegraphics[width=\linewidth]{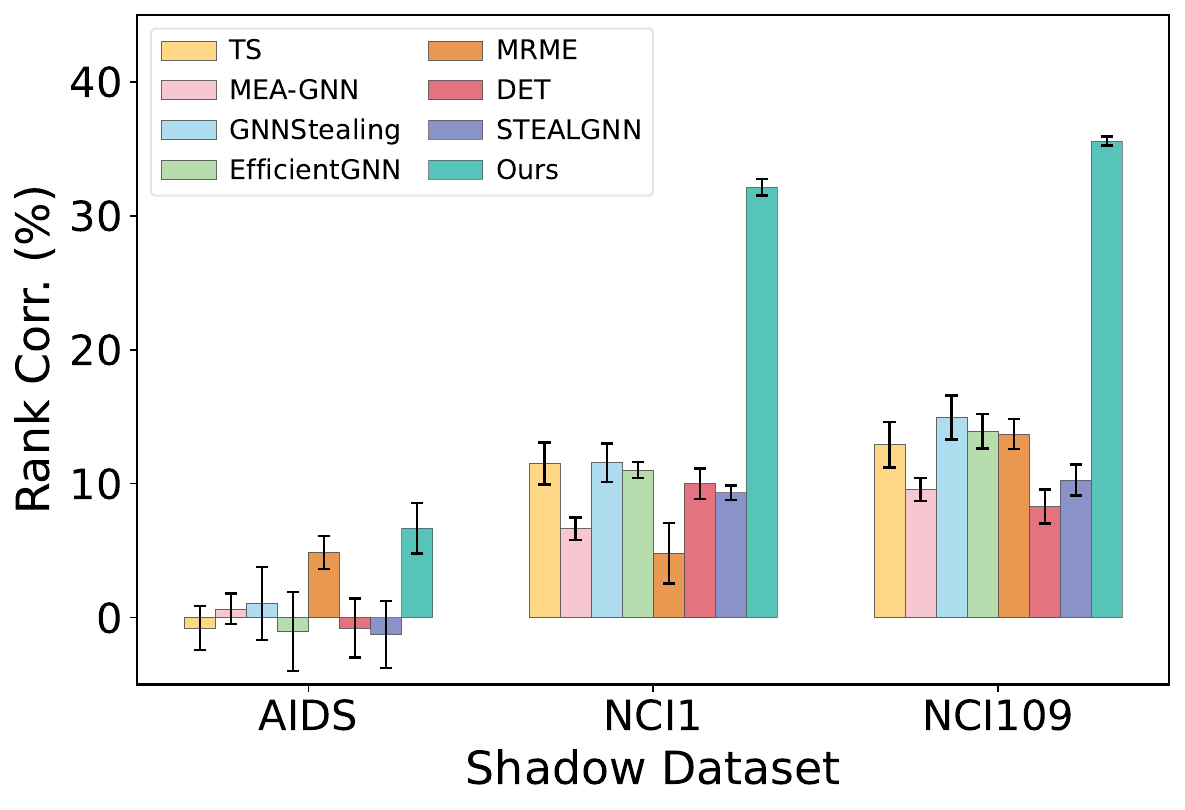}
    \caption{Rank Corr.}
    \label{fig:full_cross_dist_Rank_Corr}
  \end{subfigure}
  \vspace{-1em}
  \caption{Model extraction performance with different shadow datasets. The target model is trained on NCI109, while shadow datasets vary from in-distribution (NCI109) to out-of-distribution (NCI1 and AIDS).}
  \label{fig:full_cross_dist}
\end{figure*}

Figure~\ref{fig:full_cross_dist} presents the complete results for cross-distribution extraction performance across three metrics: AUC, fidelity, and explanation rank correlation. The results show that our explanation-guided approach maintains better performance across all metrics as distribution shift increases. The performance gap is particularly evident in explanation rank correlation, where our method achieves substantial improvements even under significant distribution shift (AIDS shadow dataset). For instance, with AIDS as the shadow dataset, our method achieves 6.68\% rank correlation while TS shows -0.77\%, and with NCI1, our method reaches 32.14\% compared to TS's 11.52\%. This suggests that explanation alignment helps capture decision-making patterns that transfer better across distributions compared to methods that rely solely on prediction mimicry.

\subsection{Additional Results on Explanation Quality Analysis}
\label{app:explanation_quality_results}

\begin{figure*}[ht]
  \centering
  \begin{subfigure}[b]{0.24\linewidth}
    \centering
    \includegraphics[width=\linewidth]{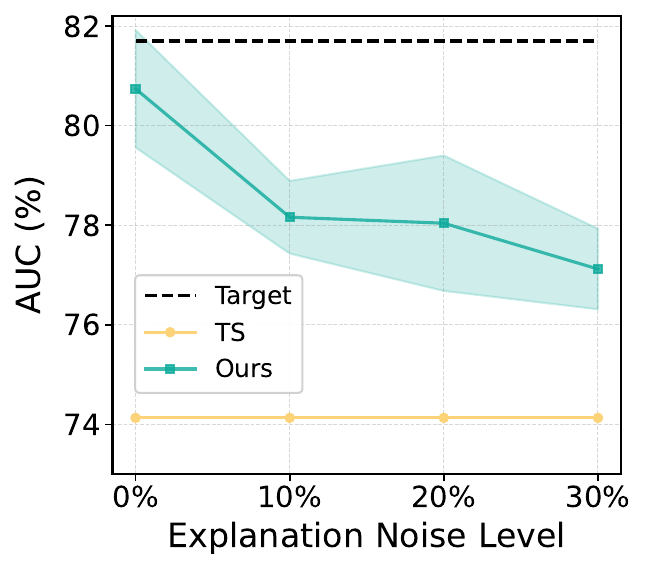}
    \caption{NCI1}
  \end{subfigure}
  \hfill
  \begin{subfigure}[b]{0.24\linewidth}
    \centering
    \includegraphics[width=\linewidth]{figs/expl_noise_NCI109_AUC.pdf}
    \caption{NCI109}
  \end{subfigure}
  \hfill
  \begin{subfigure}[b]{0.24\linewidth}
    \centering
    \includegraphics[width=\linewidth]{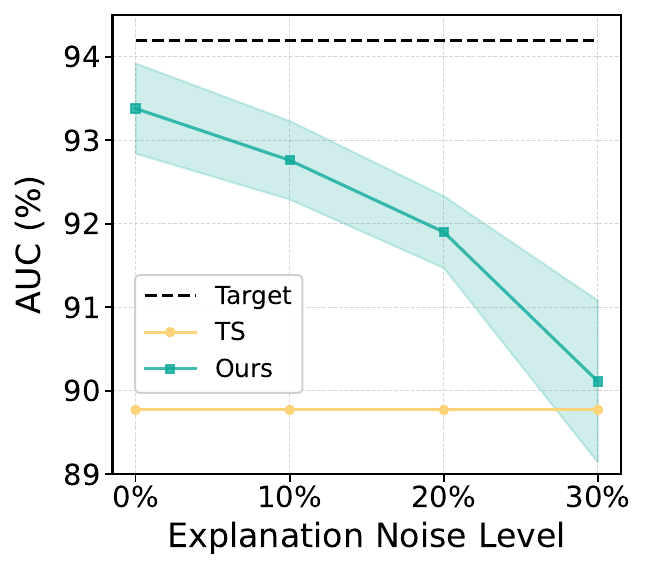}
    \caption{AIDS}
  \end{subfigure}
  \hfill
  \begin{subfigure}[b]{0.24\linewidth}
    \centering
    \includegraphics[width=\linewidth]{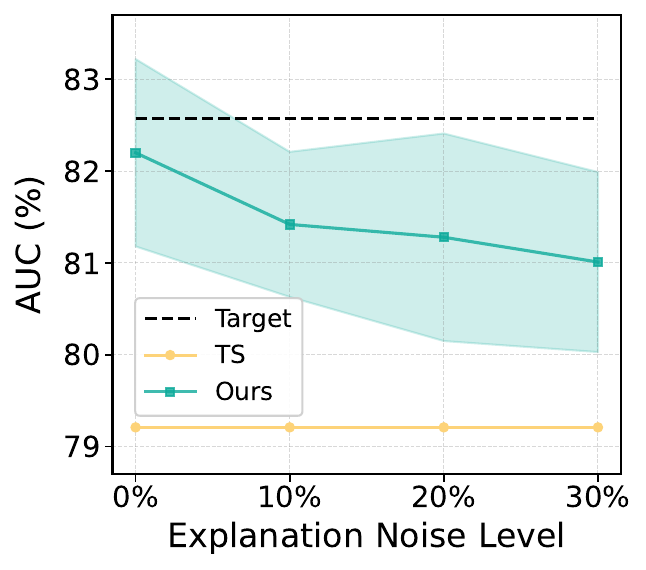}
    \caption{Mutagenicity}
  \end{subfigure}
  
  \begin{subfigure}[b]{0.24\linewidth}
    \centering
    \includegraphics[width=\linewidth]{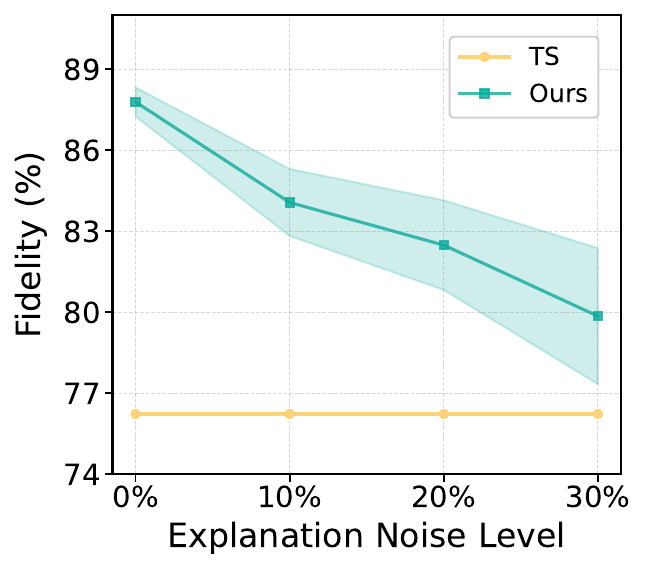}
    \caption{NCI1}
  \end{subfigure}
  \hfill
  \begin{subfigure}[b]{0.24\linewidth}
    \centering
    \includegraphics[width=\linewidth]{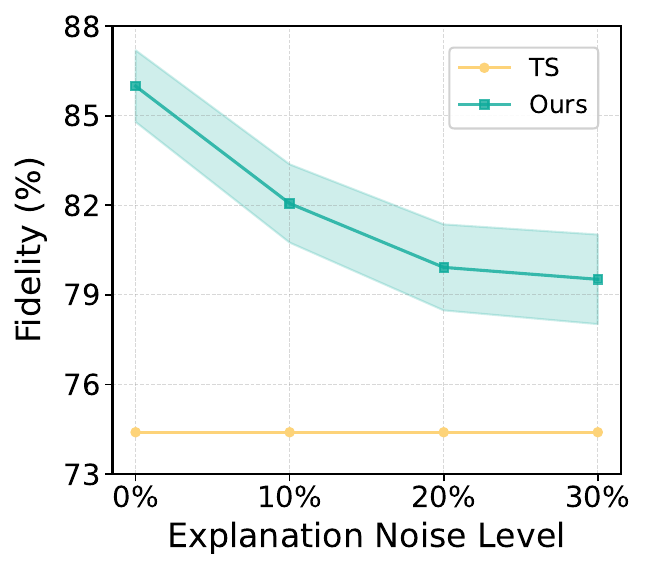}
    \caption{NCI109}
  \end{subfigure}
  \hfill
  \begin{subfigure}[b]{0.24\linewidth}
    \centering
    \includegraphics[width=\linewidth]{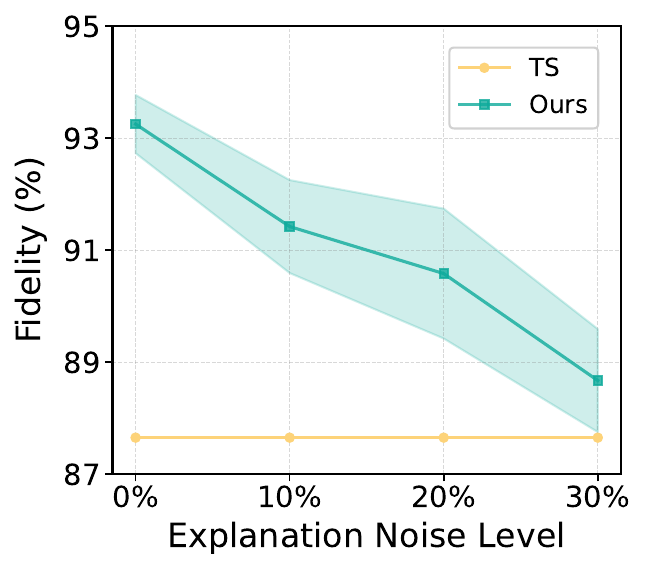}
    \caption{AIDS}
  \end{subfigure}
  \hfill
  \begin{subfigure}[b]{0.24\linewidth}
    \centering
    \includegraphics[width=\linewidth]{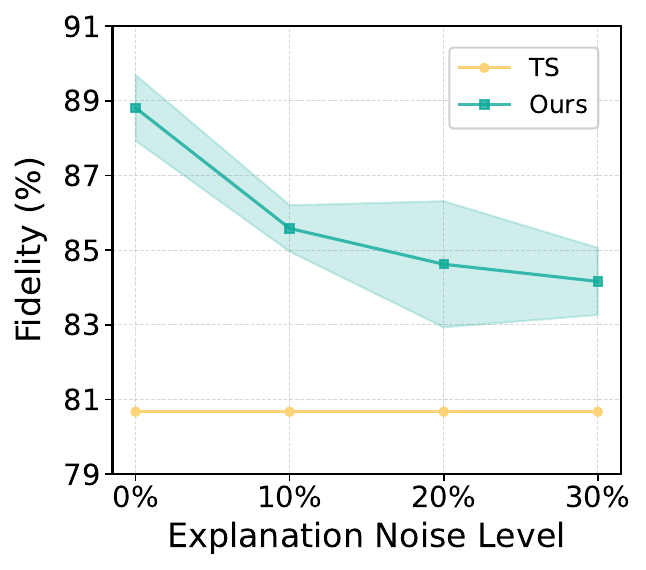}
    \caption{Mutagenicity}
  \end{subfigure}
  
  \begin{subfigure}[b]{0.24\linewidth}
    \centering
    \includegraphics[width=\linewidth]{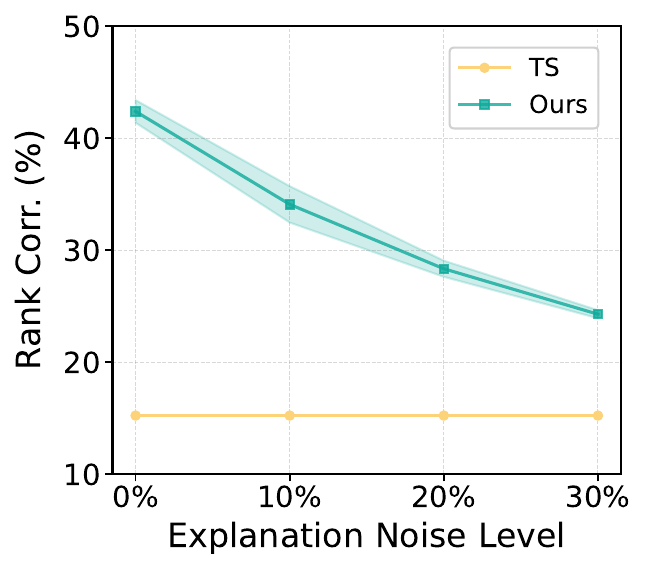}
    \caption{NCI1}
  \end{subfigure}
  \hfill
  \begin{subfigure}[b]{0.24\linewidth}
    \centering
    \includegraphics[width=\linewidth]{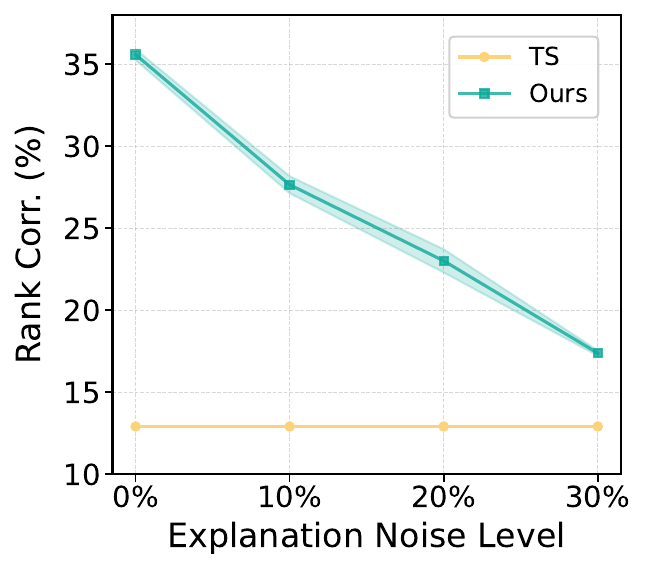}
    \caption{NCI109}
  \end{subfigure}
  \hfill
  \begin{subfigure}[b]{0.24\linewidth}
    \centering
    \includegraphics[width=\linewidth]{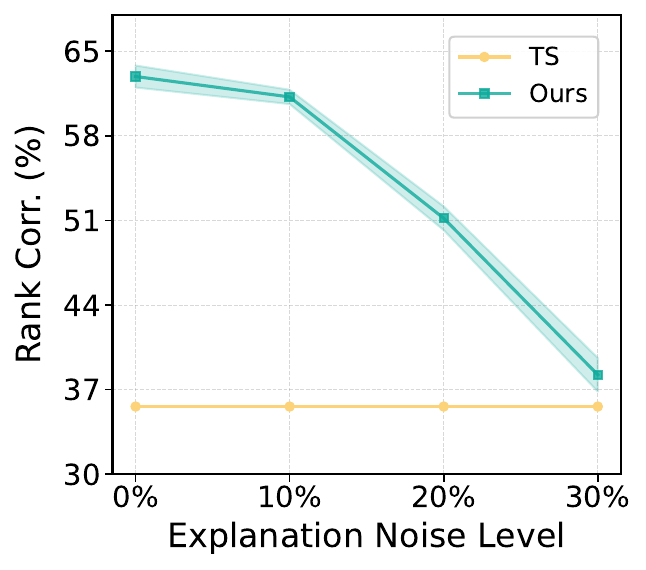}
    \caption{AIDS}
  \end{subfigure}
  \hfill
  \begin{subfigure}[b]{0.24\linewidth}
    \centering
    \includegraphics[width=\linewidth]{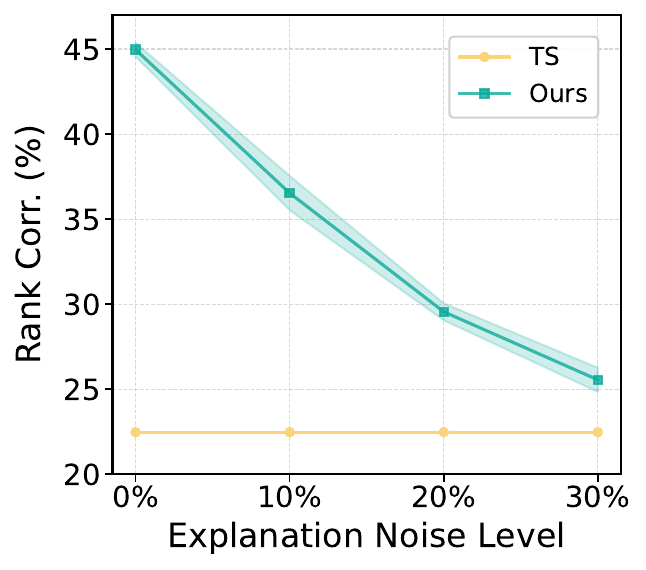}
    \caption{Mutagenicity}
  \end{subfigure}
  \vspace{-0.5em}
  \caption{Performance under different explanation noise levels across three metrics. Each plot shows our method's performance as noise increases from 0\% to 30\%, compared with TS baseline and target performance (AUC only).}
  \label{fig:explanation_noise_impact_complete}
\end{figure*}

\textbf{Complete Results for Noisy Explanations.}
Figure~\ref{fig:explanation_noise_impact_complete} presents the complete performance evaluation under different explanation noise levels across all three metrics. The results show consistent degradation patterns as noise increases from 0\% to 30\%, with our method's performance gradually declining but still maintaining improvements over the TS baseline. The explanation rank correlation metric exhibits the most significant sensitivity to noise, which is expected since our rank-based alignment mechanism relies on the relative ordering of node importance. However, even with 30\% noise injection, our method maintains meaningful improvements over the TS baseline across all datasets. The fidelity metric shows more resilience to noise compared to explanation correlation, while AUC performance demonstrates intermediate sensitivity.

\begin{figure*}[ht]
  \centering
  \begin{subfigure}[b]{0.3\linewidth}
    \centering
    \includegraphics[width=\linewidth]{figs/binary_explanation_AUC.pdf}
    \caption{AUC}
    \label{fig:binary_expl_AUC}
  \end{subfigure}
  \qquad
  \begin{subfigure}[b]{0.3\linewidth}
    \centering
    \includegraphics[width=\linewidth]{figs/binary_explanation_Fidelity.pdf}
    \caption{Fidelity}
    \label{fig:binary_expl_Fidelity}
  \end{subfigure}
  \qquad
  \begin{subfigure}[b]{0.3\linewidth}
    \centering
    \includegraphics[width=\linewidth]{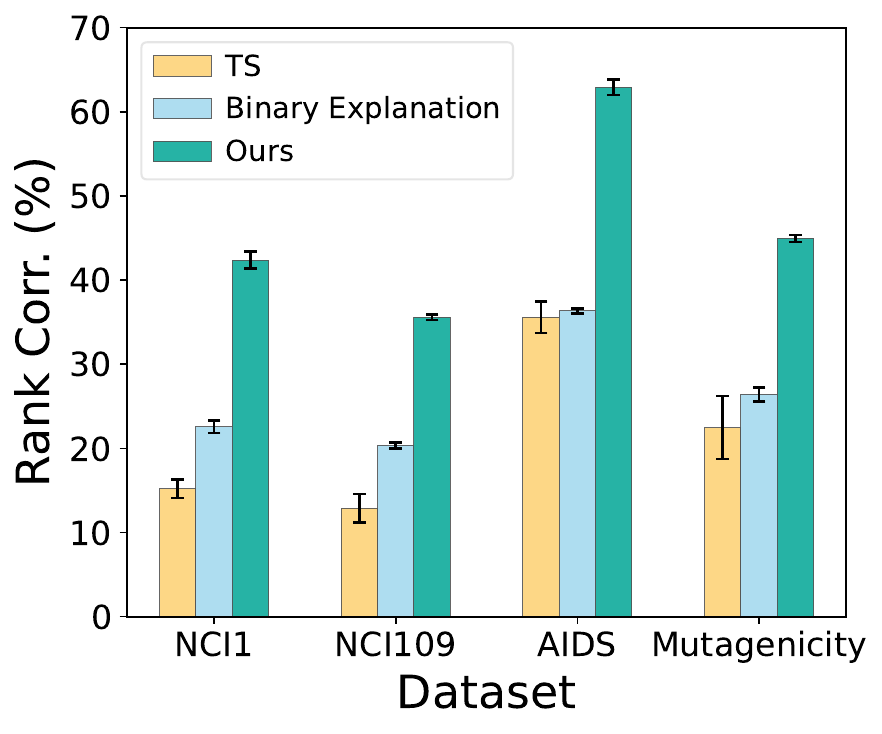}
    \caption{Rank Corr.}
    \label{fig:binary_expl_Rank_Corr}
  \end{subfigure}
  \vspace{-0.5em}
  \caption{Performance comparison with binary explanation format across datasets. Each subplot shows the performance of TS baseline, binary explanation variant, and our method on different datasets for AUC, fidelity, and explanation rank correlation metrics.}
  \label{fig:binary_explanation_complete}
\end{figure*}

\textbf{Complete Results for Binary Explanations.}
Figure~\ref{fig:binary_explanation_complete} shows detailed performance comparison between continuous and binary explanation formats across all metrics. The binary explanation setting considers scenarios where attackers can only obtain binary information distinguishing important subgraphs from irrelevant parts. Despite this significant information reduction, our method with binary explanations consistently outperforms the TS baseline across all datasets and metrics. The performance gap between our full method and the binary explanation variant varies across metrics, with explanation rank correlation showing the most pronounced reduction. This aligns with expectations that coarser explanation granularity limits the effectiveness of our rank-based alignment. Nevertheless, the binary explanation variant achieves notable improvements over the TS baseline, demonstrating the framework's adaptability to different explanation formats. The results suggest that even simplified explanation information provides valuable guidance for model extraction compared to pure prediction-based approaches.

\begin{figure*}[ht]
  \centering
  \begin{subfigure}[b]{0.3\linewidth}
    \centering
    \includegraphics[width=\linewidth]{figs/backbone_heatmap_AUC.pdf}
    \caption{AUC gains (\%)}
    \label{fig:full_backbone_auc}
  \end{subfigure}
  \qquad
  \begin{subfigure}[b]{0.3\linewidth}
    \centering
    \includegraphics[width=\linewidth]{figs/backbone_heatmap_Fidelity.pdf}
    \caption{Fidelity gains (\%)}
    \label{fig:full_backbone_fid}
  \end{subfigure}
  \qquad
  \begin{subfigure}[b]{0.3\linewidth}
    \centering
    \includegraphics[width=\linewidth]{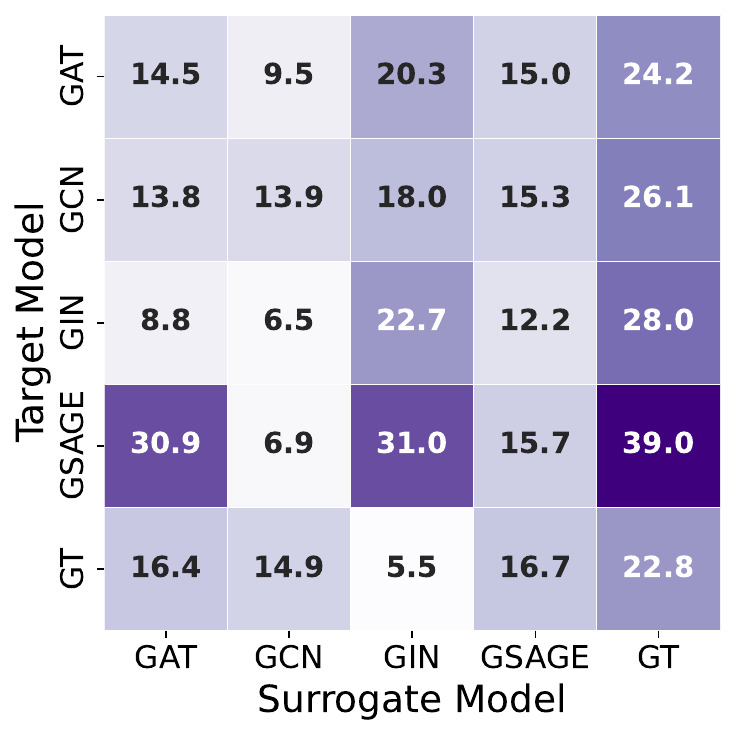}
    \caption{Rank Correlation gains (\%)}
    \label{fig:full_backbone_rank}
  \end{subfigure}
  \vspace{-0.5em}
  \caption{Performance improvements over teacher-student baseline with different model architectures on NCI109.}
  \label{fig:full_arch_analysis}
\end{figure*}

\subsection{Additional Results on the Impacts of Target and Surrogate Model Architecture}
\label{app:arch_analysis}

Figure~\ref{fig:full_arch_analysis} presents comprehensive performance gains of our explanation guided approach over the TS baseline across three metrics: AUC, fidelity, and explanation rank correlation.  The heatmaps show that our method achieves improvements across diverse target-surrogate architectural pairings, including both matched (diagonal) and mismatched (off-diagonal) configurations, which supports the generalization capability discussed in the main text. Explanation rank correlation exhibits notable gains across nearly all combinations, indicating that the explanation alignment objective can transfer decision-making patterns even when target and surrogate architectures differ. While AUC and fidelity gains show some variation across different combinations, reflecting inherent architectural differences in graph processing, the overall positive trend across all three metrics suggests that our approach is applicable to diverse GNN architectures.

\subsection{Additional Results for Different Explanation Mechanisms}
\label{app:explainer_results}

Tables~\ref{tab:explainer_nci1}, \ref{tab:explainer_aids}, and~\ref{tab:explainer_mut} present additional results for different explanation mechanisms on NCI1, AIDS, and Mutagenicity datasets, respectively. The results show that our explanation-guided approach generally achieves improvements over the TS baseline across various explainer choices, though the performance gains vary across different explainers and datasets. CAM and Grad-CAM consistently show improvements in AUC and fidelity across all datasets, while PGExplainer exhibits the most substantial gains in explanation rank correlation. It is worth noting that on the AIDS dataset, GNNExplainer and PGExplainer show slight decreases in some predictive metrics, suggesting that the effectiveness of different explainers may depend on dataset characteristics. Overall, these results indicate that our framework can work with different types of explanation mechanisms, with varying degrees of effectiveness.

\begin{table}[h]
  \caption{Performance gains of different explanation mechanisms on NCI1}
  \vspace{-0.5em}
  \label{tab:explainer_nci1}
  \centering
  \resizebox{\linewidth}{!}{
  \begin{tabular}{lccc}
    \toprule
    \multirow{1}{*}{Explainer}  
    & AUC (\%)& Fidelity (\%)& Rank Corr. (\%)\\
    \midrule
    CAM & $80.74\pm1.18$ \textcolor{gray}{\scriptsize~↑6.61}   & $87.79\pm0.54$ \textcolor{gray}{\scriptsize~↑11.56} & $42.39\pm1.00$ \textcolor{gray}{\scriptsize~↑27.27} \\
    Grad      & $76.73\pm0.88$ \textcolor{gray}{\scriptsize~↑2.60}   & $78.18\pm1.27$ \textcolor{gray}{\scriptsize~↑1.95}  & $47.02\pm1.10$ \textcolor{gray}{\scriptsize~↑44.80} \\
    Grad-CAM  & $80.39\pm0.70$ \textcolor{gray}{\scriptsize~↑6.26}   & $86.52\pm1.08$ \textcolor{gray}{\scriptsize~↑10.29} & $42.11\pm0.67$ \textcolor{gray}{\scriptsize~↑27.12} \\
    GNNExpl.  & $76.74\pm0.95$ \textcolor{gray}{\scriptsize~↑2.61}   & $78.52\pm1.33$ \textcolor{gray}{\scriptsize~↑2.29}  & $10.16\pm1.00$ \textcolor{gray}{\scriptsize~↑9.69} \\
    PGExpl.   & $76.60\pm1.89$ \textcolor{gray}{\scriptsize~↑2.47}   & $78.32\pm0.81$ \textcolor{gray}{\scriptsize~↑2.09}  & $69.10\pm0.78$ \textcolor{gray}{\scriptsize~↑69.00} \\
    \bottomrule
   \end{tabular}}
\end{table}

\begin{table}[h]
  \caption{Performance gains of different explanation mechanisms on AIDS}
  \vspace{-0.5em}
  \label{tab:explainer_aids}
  \centering
  \resizebox{\linewidth}{!}{
  \begin{tabular}{lccc}
    \toprule
    \multirow{1}{*}{Explainer} 
    & AUC (\%)& Fidelity (\%)& Rank Corr. (\%)\\
    \midrule
    CAM & $93.38\pm0.54$ \textcolor{gray}{\scriptsize~↑3.61}   & $93.25\pm0.52$ \textcolor{gray}{\scriptsize~↑5.60}   & $62.91\pm0.91$ \textcolor{gray}{\scriptsize~↑27.31} \\
    Grad      & $91.25\pm1.54$ \textcolor{gray}{\scriptsize~↑1.48}   & $88.40\pm1.71$ \textcolor{gray}{\scriptsize~↑0.75}   & $56.18\pm0.42$ \textcolor{gray}{\scriptsize~↑40.93} \\
    Grad-CAM  & $93.42\pm0.60$ \textcolor{gray}{\scriptsize~↑3.65}   & $92.15\pm1.31$ \textcolor{gray}{\scriptsize~↑4.50}   & $61.72\pm1.23$ \textcolor{gray}{\scriptsize~↑28.22} \\
    GNNExpl.  & $90.66\pm1.40$ \textcolor{gray}{\scriptsize~↑0.89}   & $86.50\pm1.39$ \textcolor{gray}{\scriptsize~$\downarrow$1.15}  & $10.78\pm0.98$ \textcolor{gray}{\scriptsize~↑10.17} \\
    PGExpl.   & $88.63\pm1.45$ \textcolor{gray}{\scriptsize~$\downarrow$1.14}  & $86.10\pm0.98$ \textcolor{gray}{\scriptsize~$\downarrow$1.55}  & $49.37\pm0.44$ \textcolor{gray}{\scriptsize~↑69.67} \\
    \bottomrule
  \end{tabular}}
\end{table}

\begin{table}[h]
  \caption{Performance gains of different explanation mechanisms on Mutagenicity}
  \vspace{-0.5em}
  \label{tab:explainer_mut}
  \centering
  \resizebox{\linewidth}{!}{
  \begin{tabular}{lccc}
    \toprule
    \multirow{1}{*}{Explainer}
    & AUC (\%)& Fidelity (\%)& Rank Corr. (\%)\\
    \midrule
    CAM & $82.20\pm1.24$ \textcolor{gray}{\scriptsize~↑2.99}   & $88.81\pm0.88$ \textcolor{gray}{\scriptsize~↑8.14}  & $44.96\pm0.41$ \textcolor{gray}{\scriptsize~↑22.49} \\
    Grad      & $79.80\pm1.12$ \textcolor{gray}{\scriptsize~↑0.59}   & $81.20\pm1.48$ \textcolor{gray}{\scriptsize~↑0.53}  & $65.16\pm0.33$ \textcolor{gray}{\scriptsize~↑53.02} \\
    Grad-CAM  & $80.15\pm0.66$ \textcolor{gray}{\scriptsize~↑0.94}   & $81.73\pm1.04$ \textcolor{gray}{\scriptsize~↑1.06}  & $22.51\pm1.26$ \textcolor{gray}{\scriptsize~↑19.83} \\
    GNNExpl.  & $81.45\pm2.04$ \textcolor{gray}{\scriptsize~↑2.24}   & $82.38\pm0.50$ \textcolor{gray}{\scriptsize~↑1.71}  & $28.02\pm0.67$ \textcolor{gray}{\scriptsize~↑26.60} \\
    PGExpl.   & $81.56\pm1.10$ \textcolor{gray}{\scriptsize~↑2.35}   & $82.58\pm1.35$ \textcolor{gray}{\scriptsize~↑1.91}  & $58.75\pm2.63$ \textcolor{gray}{\scriptsize~↑69.11} \\
    \bottomrule
  \end{tabular}}
\end{table}

\begin{figure*}[ht]
  \centering
  \begin{subfigure}[b]{0.32\linewidth}
    \centering
    \includegraphics[width=\linewidth]{figs/abl_AUC.pdf}
    \caption{Impact on AUC}
    \label{fig:abl_analysis_auc}
  \end{subfigure}
  \hfill
  \begin{subfigure}[b]{0.32\linewidth}
    \centering
    \includegraphics[width=\linewidth]{figs/abl_Fidelity.pdf}
    \caption{Impact on Fidelity}
    \label{fig:abl_analysis_fid}
  \end{subfigure}
  \hfill
  \begin{subfigure}[b]{0.32\linewidth}
    \centering
    \includegraphics[width=\linewidth]{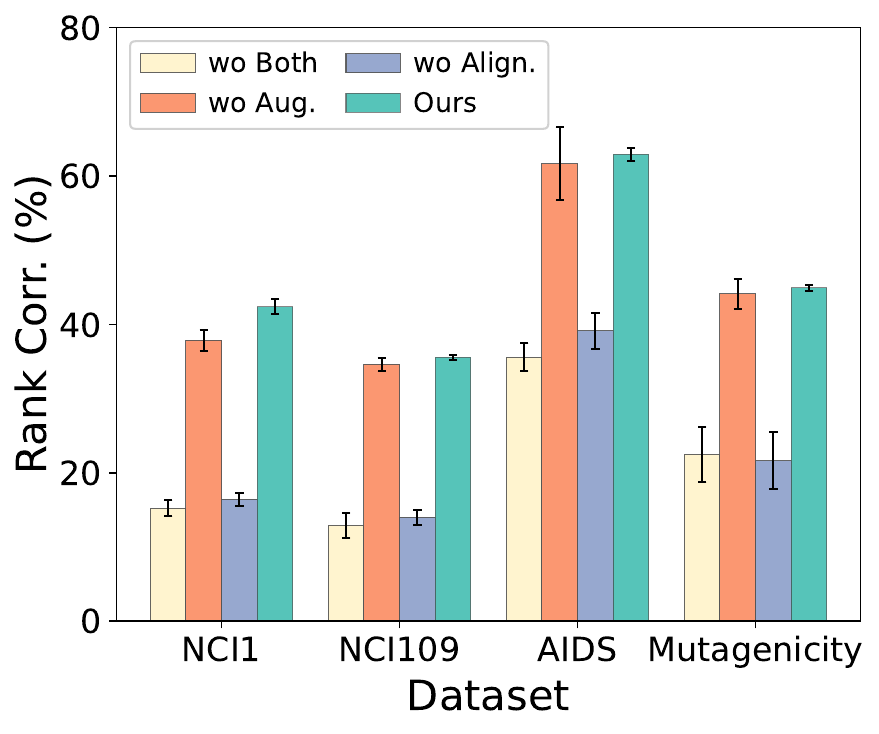}
    \caption{Impact on Rank Correlation}
    \label{fig:abl_analysis_corr}
  \end{subfigure}
  \vspace{-0.5em}
  \caption{Ablation study results across four datasets showing the impact of removing data augmentation and explanation alignment components from our full framework.}
  \label{fig:full_ablation_study}
\end{figure*}

\begin{figure*}[ht]
  \centering
  \begin{subfigure}[b]{0.31\linewidth}
    \centering
    \includegraphics[width=\linewidth]{figs/parameter_sensitivity_analysis_auc.pdf}
    \caption{Impact on AUC}
    \label{fig:param_analysis_auc}
  \end{subfigure}
  \hspace{0.01\linewidth}
  \begin{subfigure}[b]{0.31\linewidth}
    \centering
    \includegraphics[width=\linewidth]{figs/parameter_sensitivity_analysis_fid.pdf}
    \caption{Impact on fidelity}
    \label{fig:param_analysis_fid}
  \end{subfigure}
  \hspace{0.01\linewidth}
  \begin{subfigure}[b]{0.31\linewidth}
    \centering
    \includegraphics[width=\linewidth]{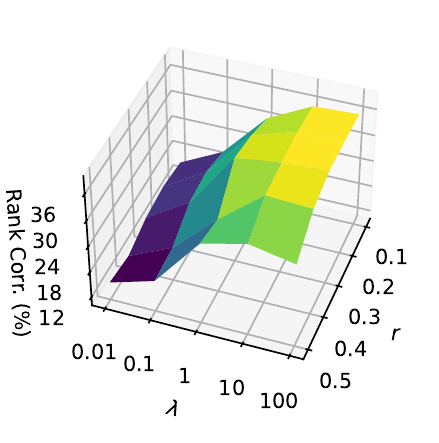}
    \caption{Impact on rank correlation}
    \label{fig:param_analysis_corr}
  \end{subfigure}
  \vspace{-0.5em}
  \caption{Sensitivity analysis of data augmentation ratio $r$ and alignment loss coefficient $\lambda$.}
  \label{fig:param_analysis}
\end{figure*}

\begin{table*}[ht]
  \caption{Training time comparison (in seconds)}
  \vspace{-0.5em}
  \label{tab:training_time}
  \centering
  \resizebox{0.8\linewidth}{!}
  {
  \begin{tabular}{lcccccccc}
    \toprule
    Dataset  & TS & MEA-GNN & GNNStealing & EfficientGNN & MRME & DET  & STEALGNN & Ours \\
    \midrule
    NCI1     & 25.87 & 26.70 & 48.81 & 55.66 & 313.48 & 60.50  & 82.43 &  109.18 \\
    NCI109   & 26.31 & 27.25 & 51.82 & 59.04 & 330.24 & 65.03 & 85.25 & 109.42 \\
    AIDS     & 10.37 & 10.40 & 21.58 & 36.79 & 180.18 & 26.85 & 45.66 & 54.58 \\
    Mutagenicity  & 28.33 & 24.84 & 55.42 &  79.08 &  308.64 & 64.75 & 81.09 & 111.75 \\
    HIV      & 75.97 & 80.64 & 105.78 & 235.40 & 1072.27 & 160.05 & 213.56& 319.08 \\
    Tox21    & 25.89 & 35.02 & 45.85 & 45.38 &  474.38 & 66.15 & 89.72& 162.77 \\
    BACE     & 11.85 & 10.62 & 19.71 & 30.29 & 141.59 & 18.66 & 29.87 & 41.72 \\
    PubMed  & 29.04 & 26.26 & 63.53 &  73.86 &  413.43 & 106.95 &  147.21 & 152.13 \\
    ogb-arxiv  & 60.72 & 59.17 & 137.27 &  144.82 &  1028.85 & 167.99 & 220.02& 435.28 \\
    \bottomrule
  \end{tabular}
  }
\end{table*}

\begin{figure*}[ht]
  \centering
  \includegraphics[width=0.95\linewidth]{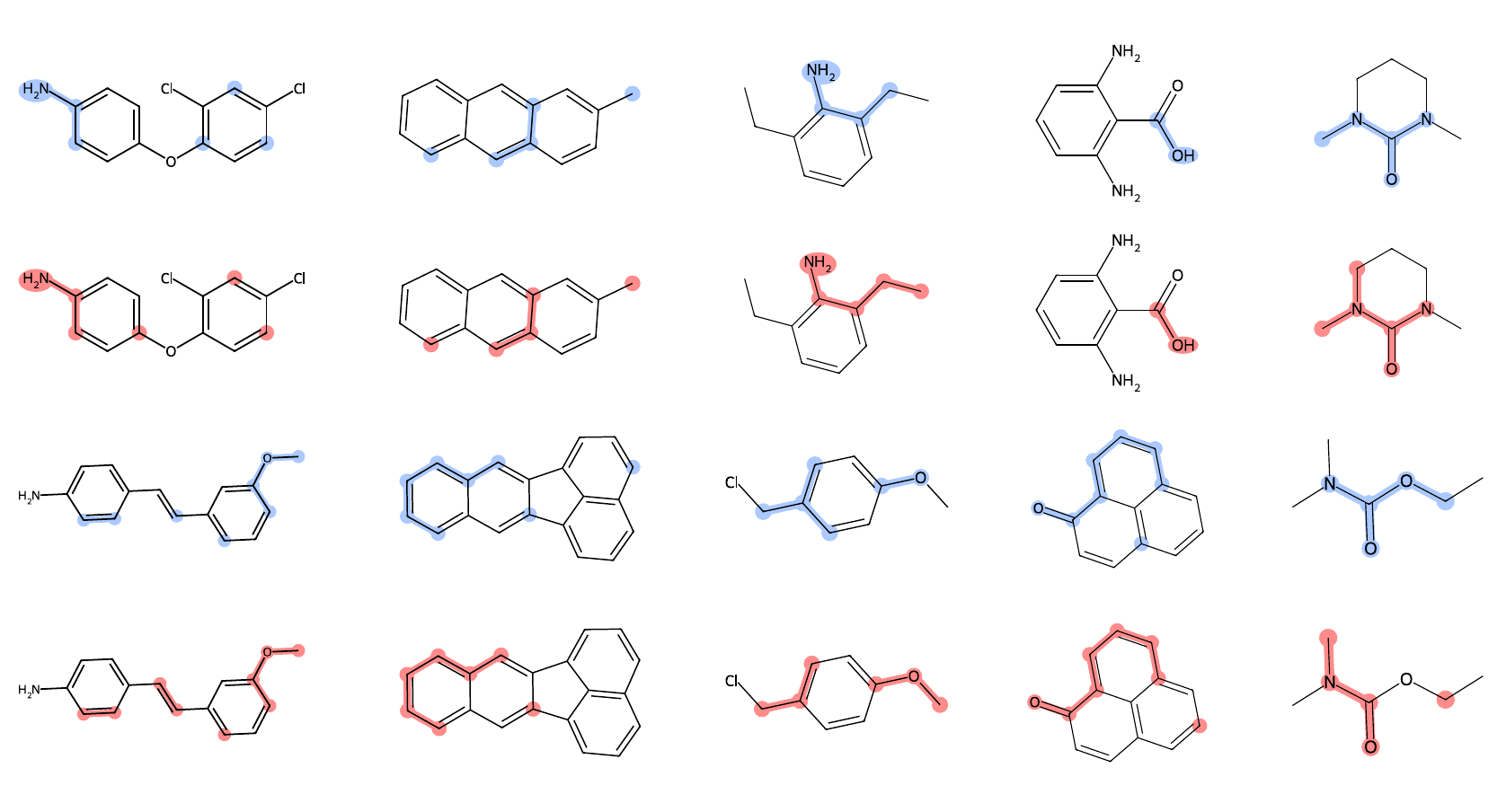}
  \caption{Visualization of molecular substructures identified as important by target and surrogate models. Blue highlights indicate target model explanations, while red highlights show surrogate model explanations.}
  \label{fig:case_study}
\end{figure*}

\subsection{Complete Ablation Study Results}
\label{app:ablation_results}

Figure~\ref{fig:full_ablation_study} presents the complete ablation study results across all metrics and datasets. The results reveal distinct contributions from each component. Removing explanation alignment ("wo Align.") causes significant degradation across all metrics, particularly in explanation rank correlation. For instance, on NCI1, rank correlation drops from 42.38\% (full method) to 16.43\% (wo Align.), while on AIDS it decreases from 62.91\% to 39.13\%. This substantial drop demonstrates that explanation alignment is the primary mechanism for capturing the target model's decision-making patterns. Predictive metrics (AUC and fidelity) also decline without alignment, though less dramatically, indicating that explanation alignment contributes to both aspects of model extraction.

Removing data augmentation ("wo Aug.") shows more modest but consistent effects. On NCI1, AUC decreases from 80.74\% to 79.56\%, and fidelity drops from 87.78\% to 83.38\%. Interestingly, rank correlation experiences a larger decline (from 42.38\% to 37.86\%), suggesting that augmented samples help the surrogate model better learn the target's explanation patterns. This pattern holds across all datasets, with data augmentation providing consistent but smaller improvements compared to explanation alignment.

When both components are removed ("wo Both"), performance matches the TS baseline across all datasets, confirming that our framework's improvements are entirely attributable to these two designs. The full method consistently achieves the best performance across all metrics and datasets, demonstrating the complementary benefits of combining data augmentation with explanation alignment.

\subsection{Hyperparameter Analysis}
\label{appendix:hyperparameter}

This section provides a comprehensive parameter sensitivity analysis that complements the AUC results presented in Section~\ref{subsection:hyperparameter}. We focus on two key hyperparameters: the data augmentation ratio $r$ (controlling the proportion of augmented samples) and the alignment loss coefficient $\lambda$ (balancing prediction matching and explanation alignment). We vary $r$ from 0.1 to 0.5 and $\lambda$ from 0.01 to 100, conducting experiments on the NCI109 dataset with settings consistent with Section~\ref{subsection:model_extraction_performance}. 

As shown in Figure~\ref{fig:param_analysis_fid}, fidelity consistently improves as alignment coefficient $\lambda$ increases, while augmentation ratio $r$ achieves optimal results between 0.1 and 0.3. Figure~\ref{fig:param_analysis_auc} reveals that neither excessively small nor large augmentation ratio $r$ yields optimal AUC performance, suggesting that a moderate amount of data augmentation is most beneficial for model stealing. Regarding the alignment of explanations illustrated in Figure~\ref{fig:param_analysis_corr}, while the data augmentation ratio $r$ demonstrates relatively consistent impact, the alignment coefficient $\lambda$ plays a crucial role in capturing the target model's decision logic. As $\lambda$ increases from 0.01 to 10, we observe significant improvement in the consistency of node importance rankings between the surrogate and target models, indicating enhanced capture of the target model's decision logic. However, when $\lambda$ exceeds a certain threshold, the improvement becomes marginal and may even compromise the surrogate model's ability to match the target model's predictions.

\subsection{Training Time}
\label{appendix:training_time}

We followed the experimental settings described in Section~\ref{subsection:model_extraction_performance} to measure the training time of the surrogate model, conducting runtime evaluations comparing EGSteal to other baseline methods across multiple datasets. As shown in the Table~\ref{tab:training_time}, although EGSteal requires more training time than some simpler baselines, its computational overhead is not the highest and remains reasonable, which is acceptable considering the performance gains achieved.

\subsection{Case Study}
\label{appendix:case_study}

We present visual analysis of molecular substructures identified as important by both target and surrogate models to demonstrate the effectiveness of our explanation alignment mechanism. Figure~\ref{fig:case_study} shows examples from the Mutagenicity dataset comparing the explanations generated by the target model and our surrogate model across different molecular structures, revealing highly similar substructure patterns between the target (blue) and surrogate (red) models. The consistent overlap between blue highlights (target model explanations) and red highlights (surrogate model explanations) across these examples confirms that our surrogate model effectively captures the target model's decision-making patterns and validates that our framework successfully replicates not only the predictive behavior but also the underlying reasoning process of the target model.

\end{document}